\documentclass[11pt]{article}

\usepackage[final]{acl}

\usepackage{times}
\usepackage{latexsym}
\usepackage{booktabs} 
\usepackage[T1]{fontenc}
\newcommand\blfootnote[1]{%
  \begingroup
  \renewcommand\thefootnote{}\footnote{#1}%
  \addtocounter{footnote}{-1}%
  \endgroup
}
\usepackage[utf8]{inputenc}
\usepackage{multirow}
\usepackage{booktabs}
\usepackage{microtype}

\usepackage{inconsolata}
\usepackage{colortbl}   
\usepackage{graphicx}
\usepackage{amsmath}
\usepackage{amssymb}
\usepackage{booktabs}       
\usepackage{tcolorbox}
\usepackage{tabularx}
\usepackage[table]{xcolor} 
\definecolor{headerblue}{RGB}{0, 102, 204} 

\usepackage[utf8]{inputenc}
\usepackage{enumitem}
\newcommand{\akash}[1]{{\color{green}[[Akash: #1]]}}

\tcbuselibrary{breakable, skins}

\title{PersuaRL: Reinforcement Learning-Driven Multi-Expert Selection for Persuasive Dialogue Generation in Insurance}

\author{
  \textbf{Rohan Kirti\textsuperscript{1*}},
  \textbf{Akash Ghosh\textsuperscript{1}},
  \textbf{Aryan Vats\textsuperscript{1}},
  \textbf{Niladri Ghosh\textsuperscript{2}},
  \textbf{Shipra Shriparn\textsuperscript{1}},
\\
  \textbf{Roshni Ramnani\textsuperscript{3}},
  \textbf{Anutosh Maitra\textsuperscript{3}},
  \textbf{Sriparna Saha\textsuperscript{1}}
\\
  \textsuperscript{1}Indian Institute of Technology Patna, India \\
  \textsuperscript{2}Ramakrishna Mission Vivekananda Educational and Research Institute, India \\
  \textsuperscript{3}Accenture Labs, Bangalore, India
}

\begin{document}
\maketitle
\blfootnote{$^{*}$Corresponding author: \texttt{rohan\_2421cs09@iitp.ac.in}}
\begin{abstract}
Large Language Models (LLMs) are revolutionizing digital communication by powering conversational agents deployed across domains such as customer service, digital sales, and insurance. These agents, built on LLMs, can understand user input, retrieve relevant information, and generate coherent responses. However, while they excel at factual communication, they often lack the ability to engage in truly persuasive, context-sensitive dialogue, 
especially in domains like insurance, where trust and clarity are critical.  Building on this need within the insurance domain, our work focuses on improving the persuasiveness of digital agents, aka LLMs. To support this, we introduce \textbf{\textit{InsureDial}}, a Persuasive \textbf{Insur}anc\textbf{e} \textbf{Dial}ogue dataset, designed to capture the nuances of persuasive communication specific to motor insurance interactions. We introduce \textbf{\textit{PersuaRL}}, a reinforcement learning based framework that equips LLM-driven dialogue agents with the ability to adaptively explore, select, and coordinate strategies across multiple expert modules, guided by the evolving dialogue context, to achieve more effective persuasion.  We conduct extensive automatic, human and qualitative evaluations on two benchmark persuasion dialogue datasets, including our \textbf{\textit{InsureDial}}.
Our evaluations consistently demonstrate that \textbf{\textit{PersuaRL}} outperforms baseline, generating contextually appropriate and highly persuasive responses. The code and dataset are available at \href{https://rohan9182.github.io/Persua-RL/}{PersuaRL}.

\end{abstract}

\section{Introduction}
Digital conversations have progressed beyond simple information exchange, with virtual dialogue agents now playing sophisticated roles in sectors like customer service, sales, and financial advising \cite{takayanagi2025generative,sree2023product,stock2022assessment}. A key emerging function in these settings is persuasion, which empowers agents not only to inform but also to strategically influence user behavior \cite{wang2019persuasion}. This need is especially critical in domains such as insurance, where agents must actively motivate users toward beneficial outcomes, like selecting appropriate policy options. Designing such persuasive agents is highly challenging: unlike standard QA systems, they must infer user intent, adapt to emotional tone, highlight policy benefits, and guide decisions constructively \cite{samad2022empathetic}. These requirements demand fine-grained reasoning and contextual awareness capabilities that often exceed the limits of monolithic LLMs, which typically lack the domain sensitivity and  nuance needed for effective persuasive engagement.

With the advancement of conversational LLMs, recent work has focused on enhancing persuasion and strategic dialogue. \cite{ramani2024persuasion} proposed a multi-agent persuasion framework where a primary agent engages users while auxiliary agents manage strategy planning and fact-checking leading to improved persuasive efficacy in simulated insurance and financial settings. Similarly, \cite{ma2025communication} introduced a multi LLM communication setup to generate persuasive dialogue data with minimal human effort, achieving fluent and strategically diverse outputs. However, most prior work emphasizes synthetic data or simulations, lacking deployment in real-world domains like motor insurance, where effective persuasion requires understanding user needs, handling queries, and guiding decisions over multi-turn interactions. Tool-augmented LLMs and VLMs   now address this gap by enabling dynamic planning and using API within coherent dialogues, surpassing static, single-turn systems \cite{shim2025tooldial,jung2025diatool,ghosh2026carepilot,halder2026arogyasutra}.


\par
\textbf{Research gap:}
While  Tool-augmented LLMs have made strides in improving reasoning capabilities through external function use \cite{wang2024empowering}; \cite{li2025torl}, they largely depend on predefined or rigid tool invocation mechanisms that limit flexibility and adaptation to task-specific objectives. Recent works like \cite{schick2023toolformer} and \cite{lu2025octotools} introduce flexible tool use, but they focus on factual tasks, not strategic persuasion.
Meanwhile, studies on persuasive dialogue \cite{breum2024persuasive}; \cite{jin2024persuading}; \cite{karinshak2023working} have shown that LLMs can influence opinions using social-pragmatic strategies. However, these works still face limitations in generalizing tool-use policies to novel tools aligned with persuasive intent, and often struggle to maintain context-sensitive strategy selection in extended multi-turn dialogues.


\textbf{Motivation:} We reconceptualize expert coordination for persuasive dialogue as a context-conditioned decision problem, where selecting the right combination of experts becomes a learnable action rather than a heuristic choice. 
Unlike prior approaches based on static prompting, our framework formulates expert selection as an explicit action space and trains the selector and generator by alternating optimization: the selector is updated with reinforcement learning against a temporarily fixed generator, and the generator is then fine-tuned on the highest-reward expert combinations the selector discovers. Over training the two components co-adapt, allowing them to jointly discover coordination patterns that cannot be specified a priori.
To ground this framework in a realistic domain, we introduce \textit{\textbf{InsureDial}}, a curated motor insurance dialogue dataset annotated with user intent, sentiment, persuasion strategy, and key domain terms. Building on these annotations, we propose \textit{\textbf{PersuaRL}}, a modular expert-based framework comprising (i) a lightweight Selector trained via Group Relative Policy Optimization (GRPO) to dynamically choose relevant experts per turn, (ii) specialized Experts capturing core persuasion competencies, and (iii) a Generator, fine-tuned on selector-chosen expert signals, that integrates them into coherent and persuasive responses. A composite reward function guides the selector to jointly optimize strategy alignment, intent consistency, contextual relevance, and response diversity at each turn of a multi-turn dialogue.

\textbf{Contributions: }The main contributions of the paper are:

\textbf{(i)}\textbf{ Framework.} We propose {\it \textbf{PersuaRL}}, a reinforcement learning-based multi-expert framework for persuasive dialogue generation in the insurance domain, where a policy selector dynamically coordinates between task-specific experts to generate persuasive, context-aware responses. \par
\textbf{(ii)}\textbf{ Benchmark.} We introduce {\em {\textbf{InsureDial}}}, a novel and high-quality dialogue dataset for motor insurance, annotated across four dimensions like intent, sentiment, key terms, and engagement strategy, constructed through a hybrid human-in-the-loop and LLM generation pipeline. \par
\textbf{(iii)}\textbf{ Reward Design.} We develop domain-specific \textbf{reward design} incorporating persuasion strategy alignment, intent consistency, contextual coherence, response diversity, and a judge based reward, enabling effective RL-based expert selection without requiring intermediate supervision. \par

(iv)\textbf{ Evaluation.} To demonstrate the effectiveness of our framework, we conduct comprehensive automated, human, and qualitative evaluations across both in-domain (our insurance dataset, \textbf{InsureDial}) and out-of-domain (\textbf{tourism}) benchmarks. The results show that \textbf{\textit{PersuaRL}} achieves strong performance on our benchmark while also exhibiting robust generalization to out-of-domain datasets.





\section{Related Works}

\textbf{Tool-Augmented LLM:} Early work on tool use in LLMs focused on fixed, trigger-based systems for tasks like math reasoning \cite{wang2024empowering,jin2024persuading,yue2023mammoth,chen2022program}, but lacked flexibility. Iterative and supervised tool-use mechanisms followed \cite{wang2023mathcoder,chen2025empirical}. Recent advances treat tool use as a learnable policy: \cite{li2025torl} employed tree-structured RL for strategic invocation, while \cite{singh2025agentic} learned reward-based policies without intermediate supervision. Contrasting scalable paradigms include self-supervised Toolformer \cite{schick2023toolformer} and training-free modular OctoTools \cite{lu2025octotools}, highlighting a shift toward dynamic, efficient tool selection.

\textbf{Persuasion Support Conversations:}
Recent studies have probed LLMs' persuasive abilities in dialogue. \cite{breum2024persuasive} showed LLMs can influence opinions using social-pragmatic cues, while \cite{karinshak2023working} found GPT-3’s public health messages often surpassed official content in impact. \cite{jin2024persuading} proposed a multi-domain persuasive dialogue dataset and an intent-to-strategy generation model. Persona grounding was shown to enhance coherence and engagement in persuasive exchanges \cite{zhang2018personalizing}. \cite{costello2024durably} found GPT-4 Turbo could durably reduce conspiracy beliefs through evidence-based dialogues. In negotiation tasks, \cite{bianchi2024well} observed LLMs effectively deploying assertive strategies in multi-turn settings.  \citet{kirti2026can} benchmarked LLMs and VLMs on persuasion datasets and found persuasiveness varies substantially with prompting strategy and backbone, while \citet{bozdag2026must} survey computational persuasion and highlight reliable evaluation of persuasiveness as an open challenge.
In task-oriented sales settings, persona-aware persuasive dialogue
policies have been proposed to handle goal unavailability by persuading users toward servable alternatives \citep{tiwari2022newpersona, tiwari2023towards}. \citet{raut2022introducing} introduced multi-modality into persuasive
task-oriented sales agents. \par


\textit{Unlike prior persuasive dialogue systems that rely on static prompting, heuristic routing, or monolithic generation, this work formulates expert coordination for persuasion as a learnable, context-conditioned decision problem. Crucially, we train the selector and generator by alternating optimization, so that the two components iteratively adapt to each other over the course of training.
\textbf{\textit{PersuaRL}} is one of the earliest framework to learn expert selection policies for persuasive dialogue via reinforcement learning, with coordination that emerges from this co-adaptive training rather than being pre-specified.}

\section{Development of \textbf{\textit{InsureDial}} Dataset}
Prior to this work, no persuasive dialogue datasets existed for the motor insurance domain. To address this gap, we introduce \textbf{\textit{InsureDial}}, a domain-specific dataset designed for persuasive motor insurance conversations. It contains 1,931 multi-turn conversations (26,000+ utterances) between users and agents, focusing on both information delivery and strategic persuasion. Scenarios span real-world intents like policy quotes, coverage details, and price inquiries, with rich domain terminology. 
The dataset was developed using a semi-automated, human-in-the-loop generation framework, where GPT-4o \cite{openai2025gpt4_o} was employed to draft diverse persuasive dialogue scenarios, and human annotators subsequently reviewed and filtered the conversations to ensure linguistic quality and domain accuracy.



\subsection{Dataset Preparation}
To build \textbf{\textit{InsureDial}}, we analyzed leading motor insurance websites to extract real-world terminology and interaction flows, covering categories like \textit{Discounts} (e.g., No Claim Bonus), \textit{Value-Added Services} (e.g., Roadside Assistance), \textit{Coverage Types}, and \textit{Add-ons}. 
The statistics of the dataset are mentioned in Table \ref{statistics} .

%

\subsubsection{Drafting Seed Dialogues}

We began by creating 50 high-quality seed dialogues entirely authored by trained human annotators simulating persuasive motor insurance conversations. In each dialogue, one annotator played the user while the other acted as the agent \cite{kelley1984iterative}, aiming to guide decisions through strategic and informative responses. The seed dialogues were deliberately designed to cover a comprehensive range of persuasive intents and their possible combinations within the motor insurance domain. Agent utterances were validated with domain experts to ensure factual accuracy and persuasive quality. Annotators\footnote{Annotators were fairly compensated at rates of \$10/hour for utterance task and \$5/hour for verification.} had postgraduate training and linguistic experience, ensuring natural and strategic interactions. These dialogues served as prompt exemplars for large-scale LLM-assisted generation in subsequent stages.


\subsubsection{Generating Dialogues.}
To identify the most effective prompting strategy, we began with four seed dialogues and experimented with five carefully designed prompts. \textit{The prompt is shown in Appendix \ref{sec:Prompts}.} 
Using the GPT-4o model (temperature = 0.8, top‑$p$ = 0.95), we generated 25 dialogues for each prompt to promote diverse yet coherent outputs. These were then manually evaluated by expert annotators on a three-point scale (1 = low, 2 = moderate, 3 = high) focusing on persuasive quality. The process achieved a substantial inter-annotator agreement (Kappa = 80.2\%) \cite{mchugh2012interrater}, confirming rating consistency. The prompt that yielded the highest number of high-scoring (score = 3) dialogues was selected to guide the full-scale generation of the \textbf{\textit{InsureDial}} dataset, ensuring that the final collection maintained both persuasive strength and domain relevance.
\begin{table}[h]
\resizebox{\columnwidth}{!}{%
\begin{tabular}{|l|c|c|c|}
\hline
\textbf{Metric} & \textbf{Train} & \textbf{Validation} & \textbf{Test} \\ \hline
\textit{Number of Dialogues} & 1545 & 97 & 289 \\ \hline
\textit{Number of Utterances} & 21134 & 1462 & 4170 \\ \hline
\textit{Avg Utterances per Dialogue} & 6.84 & 7.54 & 7.21 \\ \hline
\textit{Avg Words per User Utterance} & 14.38 & 17.21 & 17.19 \\ \hline
\textit{Avg Words per Agent Utterance} & 53.78 & 59.74 & 56.94 \\ \hline
\end{tabular}%
}
\caption{Dataset statistics of \textsc{InsureDial}}
\label{statistics}
\end{table}


\subsubsection{Dataset Annotation}
The \textbf{\textit{InsureDial}} dataset was annotated across four key dimensions to enable the development of effective persuasive dialogue systems: persuasion strategy, key domain terms, intent, and sentiment. Persuasion strategies were assigned to agent utterances to capture the underlying tactics used to influence user decisions. Domain-specific key terms were tagged to ensure accurate and context-aware responses. User turns were labeled with intents to reflect the communicative goal of each utterance, while sentiment was annotated to capture emotional tone and guide adaptive persuasive strategies. Annotation was conducted using a hybrid approach, where Gemini-2.0-Flash \cite{google_gemini_2_0_flash} provided initial labels that were then verified and refined by human annotators, ensuring high-quality and consistent annotations for all dialogues. 
\textit{The full details of the dataset construction and annotations are given in Appendix \ref{sec:Dataset_Annotation}.
}
\section{Methodology}

\textbf{Problem Formulation:}
Given a dialogue context \(x_t\), comprising the conversation history and the current user utterance, the goal is to generate a persuasive response \(y_t\). A selector policy \(\pi_\theta\) chooses a subset of expert modules via a binary mask \(o_t \in \{0,1\}^n\), where \(o_{t,i}=1\) activates expert \(T_i\). Each selected expert produces an output \(O_i = T_i(x_t)\), which is combined with the dialogue context to form an augmented prompt
\begin{equation}
U(x_t, o_t) = \mathrm{Pack}\!\left(x_t;\{O_i \mid o_{t,i}=1\}\right).
\end{equation}
A generator \(A_\phi\) then produces the final response \(y_t = A_\phi(U(x_t, o_t))\). 
At each dialogue turn, the dialogue context is treated as the state \(s_t \triangleq x_t\), and the expert-selection mask \(o_t\) as the action. During training the dialogue history follows the gold dialogue, so the selection at turn $t$ does not alter the context at turn $t+1$; the selector therefore solves a context-conditioned, single-step decision problem at every turn, with reward \(r_t = R(y_t, x_t)\) computed on that turn's response.
The selector and generator are trained by alternating optimization. In the selector step, the generator is held fixed and the selector is updated with Group Relative Policy Optimization (GRPO) from the rewards of sampled expert selections; in the generator step, the generator is fine-tuned on the expert-augmented input of the highest-reward selection. Holding the generator fixed within the selector step keeps the reward signal stationary, while alternating the two steps lets the components co-adapt across training.

\subsection{Selector Module} 
\label{sec:Selector_Module}
Persuasive dialogue requires balancing multiple competing objectives, including engagement strategy, intent alignment, coherence, and diversity. Existing tool-use RL methods \cite{li2025torl,singh2025agentic} focus on factual tasks with objective correctness signals, whereas persuasion involves subjective, interacting rewards that demand fundamentally different optimization strategies. In this work, for context-aware persuasive dialogue generation, the \textit{Selector} learns a context-conditioned policy $\pi_\theta$ that selects a subset of expert modules at each dialogue turn.
Given dialogue context $x_t$, the policy outputs a binary selection vector $o_t \in \{0,1\}^n$, where $o_{t,i}=1$ indicates activation of expert $T_i$.
Expert selection is treated as a reinforcement learning problem, where rewards are obtained from the quality of the generated response.
We optimize the Selector using GRPO. For each state, $G = 8$ expert selections are sampled from the previous policy and each is decoded by the current generator, which is held fixed for the duration of the selector step so that all G candidates are scored against a stationary generator; the policy is then updated using a clipped policy-gradient objective with KL regularization:
\begin{equation}
\begin{aligned}
J_{\text{Selector}}(\theta) = \mathbb{E}_{s \sim \mathcal{D}} & \mathbb{E}_{\{o_j\}_{j=1}^{G} \sim \pi_{\theta_{\text{old}}}} \bigg[ \\
& \frac{1}{G} \sum_{j=1}^{G} \min \Big( r^{\text{ratio}}_j A_j, \\
& \text{clip}(r^{\text{ratio}}_j, 1-\epsilon, 1+\epsilon) A_j \Big) \\
& - \beta D_{\text{KL}} \big( \pi_\theta \,\|\, \pi_{\text{ref}} \big) \bigg]
\end{aligned}
\end{equation}

where $r^{\text{ratio}}_j = \frac{\pi_\theta(o_j \mid s)}{\pi_{\theta_{\text{old}}}(o_j \mid s)}$.
This objective enables stable learning over the combinatorial expert-selection space.

\subsection{Experts Module}
Persuasive communication is inherently multi-faceted, requiring simultaneous understanding of what the user wants (intent), how they feel (sentiment), what they're discussing (keyterms), and how best to engage them (strategy) . With this motivation in this work, each expert module $T_i$ functions as a task-specific tool, instantiated as a transformer-based decoder-only model fine-tuned independently for a particular subtask. Given the dialogue context $x$, each expert generates an output $O_i = T_i(x)$, capturing a distinct aspect of dialogue understanding or persuasive reasoning. The expert modules used in this work are: 

\begin{enumerate}

\item \textbf{Engagement Expert: }
The Engagement Expert identifies the most contextually appropriate persuasion strategy \cite{tiwari2022persona} for the current turn. It is finetuned on labeled data with six persuasion strategies. 

\item \textbf{Intent Expert:} The Intent Expert classifies the user’s intent at each turn, which is critical for adapting the persuasion strategy. To optimize the model, an NLL loss is used. 

\item \textbf{Keyterm Expert:} 
The Keyterm Expert identifies critical domain-specific terms relevant to the user's utterance (e.g., “depreciation,” “roadside assistance”). 
The output is a structured list of important keyphrases $O_{keyterm}$. 

\item \textbf{Sentiment Expert:} 
The Sentiment Expert is responsible for detecting the user’s emotional tone, categorized into three classes: $\{ \text{Positive, Neutral, Negative} \}$.  It is trained using a cross-entropy classification loss. 

\end{enumerate}
\textit{A detailed description of the expert module is provided in Appendix~\ref{sec:Expert_Module}.
}

\subsection{Generator Module}

\begin{figure*}[t]
\centering
\includegraphics[width=\textwidth]{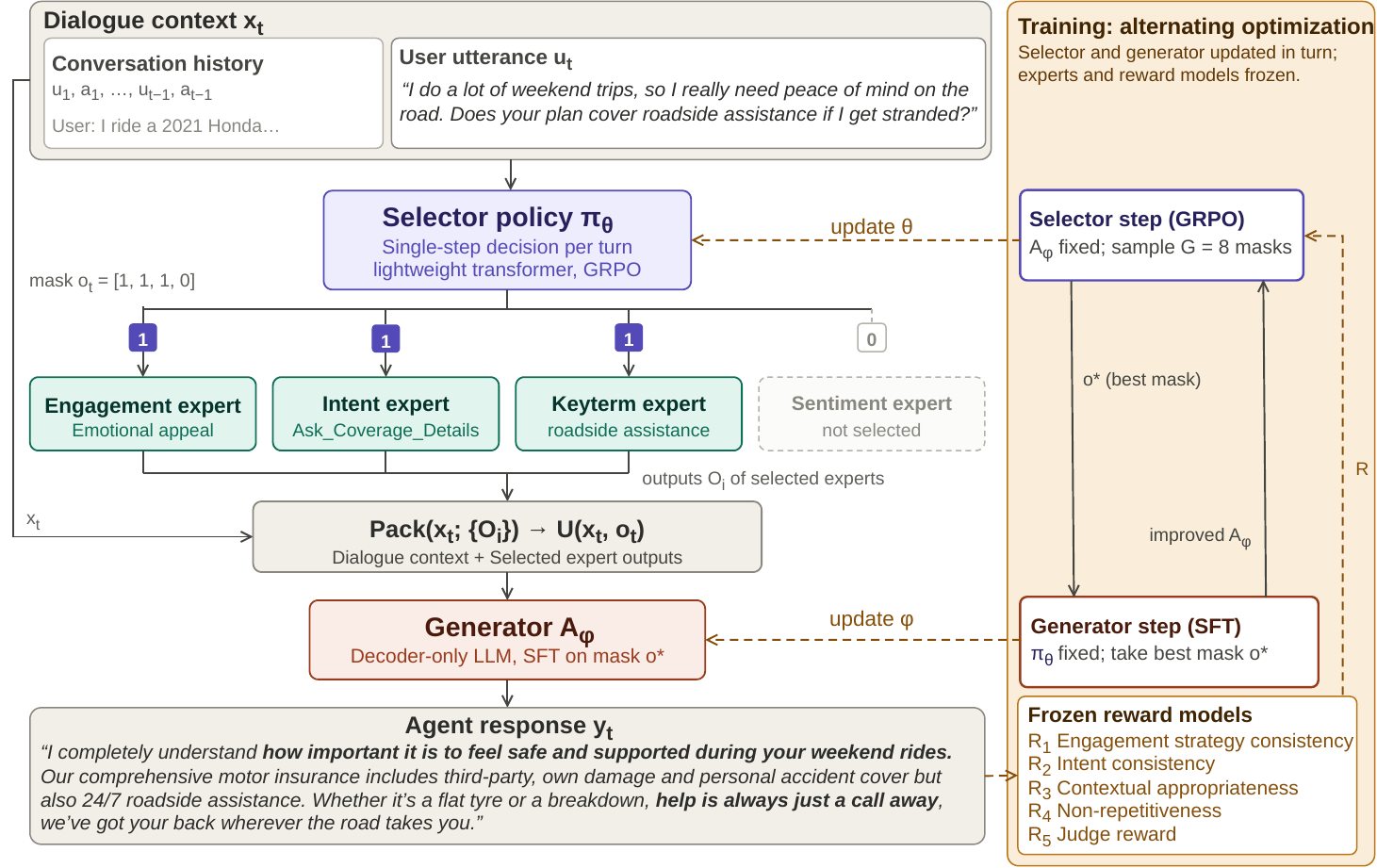} 
\caption{\textbf{\textit{PersuaRL}} architecture with selector, generator, and reward-guided response generation. Selector and generator are trained by alternating optimization; experts and reward models are frozen. Solid arrows mark the inference path followed at every turn, dashed arrows carry training-time signals only, and the faded box denotes an expert not selected by the mask.}
\label{Architecture}
\end{figure*}



\noindent The generator serves as the response generator that conditions on both the user input and the outputs of the selected experts. Given a dialogue input \(x\) and the expert set \(S=\{i \mid o^{*}_i=1\}\) chosen by the selector, each expert produces text \(O_i(x)\), and we build a fused context \(U(x,o^{*})=\mathrm{Pack}\!\left(x;\{O_i(x)\}_{i\in S}\right)\), where \(\mathrm{Pack}(\cdot)\) denotes concatenation with separators (and truncation to the context window). A decoder-only generator \(A_\phi\) conditions on \(U(x,o^{*})\) and produces the response token-by-token with \(p_\phi(y_t \mid y_{<t}, U(x,o^{*})) = A_\phi(y_{<t}, U(x,o^{*}))\), and we fine-tune \(A_\phi\) with the negative log-likelihood objective \(L_{\text{gen}}(\phi) = -\sum_{t=1}^{T} \log p_\phi(y_t \mid y_{<t}, U(x,o^{*}))\), where \(y=(y_1,\dots,y_T)\) is the ground-truth response and \(o^{*}\) is the highest-reward expert selection from the preceding selector step. This generator step alternates with the selector step of Section~\ref{sec:Selector_Module}: the generator learns to exploit the expert combinations the selector discovers, and the selector's subsequent rollouts are scored by the improved generator.


\subsection{\textbf{Rewards}}
Persuasive effectiveness cannot be captured by a single metric. A response might be strategically appropriate but miss the user's intent, or address intent accurately while lacking emotional resonance. Prior work on persuasive dialogue \cite{breum2024persuasive,jin2024persuading} evaluates systems using post-hoc human judgments, but does not incorporate these quality dimensions during training. To overcome this limitation and consistent with evidence that reward models can be designed to generalize beyond their training distribution and better postraining \cite{ghosal2025relic,ghosh2026rado,onyame2026cure}, we propose a composite reward function that provides explicit training signals for five dimensions of persuasive quality, enabling the selector to learn expert coordination patterns that balance competing objectives. The proposed rewards are: 

\textbf{Engagement Strategy Consistency Reward (R1):} 
To ensure consistency with the user’s persuasion intent, we introduce a strategy alignment reward that encourages generated responses to be semantically aligned with the user’s inferred persuasion strategy, estimated via a BERT-based classifier \cite{devlin2019bert}.

\textbf{Intent Consistency Reward (R2):} 
To ensure consistency with the user’s intent, we introduce an intent alignment reward that encourages generated responses to be semantically aligned with the user’s inferred intent, estimated via a BERT-based classifier. 

\textbf{Contextual Appropriateness Reward (R3): }To encourage contextual relevance, we introduce a reward that promotes semantic alignment between the generated response, the full dialogue context, and the current user utterance, with greater emphasis on the user’s latest turn. The reward is computed using a semantic similarity metric. 

\textbf{Non-Repetitiveness Reward (R4): }
To encourage response diversity and reduce redundancy across dialogue turns, we introduce a non-repetitiveness reward that penalizes lexical overlap between the current generated response and the previous model response. This reward helps prevent repetitive or stagnant dialogue behavior. 

\textbf{Judge Reward (R5): }To capture high-level persuasive quality beyond surface-level signals, we introduce a judge-based reward that evaluates generated responses along dimensions such as persuasiveness, negotiation effectiveness, and user engagement, using an LLM acting as an automatic evaluator.\textit{ The judge prompt is given in section \ref{sec:Prompts}.} 

The overall reward is computed as: \( R = \beta_1 R_1 + \beta_2 R_2 + \beta_3 R_3 + \beta_4 R_4 + \beta_5 R_5,\ \text{where } \beta_1 + \beta_2 + \beta_3 + \beta_4 + \beta_5 = 1. \) In addition to the rewards, we employed auxiliary penalty terms to promote efficient, diverse, and balanced expert selection during training.
\textit{We provide a detailed description of all rewards and penalties in the Appendix \ref{sec:penalties}.}

\section{Experimental Results and Analysis}
This section presents the experimental setup and provides a thorough evaluation of the proposed model, \textit{\textbf{PersuaRL}}, through automatic, human, and qualitative assessments.



\subsection{Data Preprocessing}

We use \textbf{\textit{InsureDial}}, curated via a semi-automated pipeline combining LLM generation with human refinement. Dialogues cover diverse motor-insurance scenarios and persuasion strategies, segmented into user–agent turns to preserve multi-turn context. Annotations (intent, sentiment, key terms, and persuasion strategy) were quality-checked for consistency with dialogue flow. The corpus is split 80/5/15 into train/validation/test, with diversity maintained across splits. 
\textit{Additional experimental setup are shown in \ref{sec:Experimental_Setup}.}

\begin{table*}[t]
\centering
\footnotesize
\begin{minipage}{\textwidth}
\centering
\resizebox{\textwidth}{!}{%
\begin{tabular}{llcccccc}
\toprule
\textbf{Dataset} & \textbf{Models} & \textbf{BLEU-2} $\uparrow$ & \textbf{METEOR} $\uparrow$ &
\textbf{BERTF1} $\uparrow$ & \textbf{DISTINCT-2} $\uparrow$ & \textbf{ROUGE-1} $\uparrow$ & \textbf{LLM-J} $\uparrow$ \\
\toprule
\multirow{19}{*}{\textbf{InsureDial}}
& GPT 5                             & 0.036 & 0.093 & 0.828 & 0.982 & 0.232 & --\\
& GPT 4.1 mini                      & 0.124 & 0.143 & 0.620 & 0.998 & 0.383 &--\\
& Deepseek R1 Distill Llama 70B     & 0.069 & 0.125 & 0.569 & 0.920 & 0.260 & 4.25\\
& Llama 3.3 70B Instruct            & 0.126 & 0.137 & 0.610 & 0.996 & 0.377 &4.16\\
& Qwen 3 32B                        & 0.107 & 0.132 & 0.588 & 0.998 & 0.371 &3.84\\
& Phi-3-Medium 14B                  & 0.169 & 0.167 & 0.655 & 0.995 & 0.441 &3.78\\
& Qwen 2.5 7B instruct              & 0.124 & 0.145 & 0.604 & 0.958 & 0.385 &3.66\\
& Llama 3.1 8B instruct             & 0.132 & 0.146 & 0.605 & 0.966 & 0.388 &3.67\\

\cmidrule(lr){2-8}
& Qwen 2.5 3B Instruct (Single)     & 0.090 & 0.128 & 0.562 & 0.965 & 0.310 & 2.66\\
& Qwen 2.5 3B Instruct (SFT)        & 0.305 & 0.217 & 0.727 & 0.991 & 0.556 & 3.28\\
& \textbf{PersuaRL (Qwen 2.5 3B)}   & \textbf{0.375} & \textbf{0.250} & \textbf{0.760} & \textbf{0.991} & \textbf{0.609} & \textbf{3.81}\\

\cmidrule(lr){2-8}
& Llama 3.2 3B Instruct (Single)    & 0.106 & 0.135 & 0.585 & 0.937 & 0.334 & 2.86\\
& Llama 3.2 3B Instruct (SFT)       & 0.339 & 0.232 & 0.742 & 0.989 & 0.584 & 3.48\\
& \textbf{PersuaRL (Llama 3.2 3B)}  & \textbf{0.398} & \textbf{0.276} & \textbf{0.771} & \textbf{0.989} & \textbf{0.631} & \textbf{3.95}\\
\cmidrule(lr){2-8}
& Phi 3 mini 128k (Single)          & 0.181 & 0.156 & 0.641 & 0.980 & 0.429 & 2.79\\
& Phi 3 mini 128k (SFT)             & 0.362 & 0.242 & 0.752 & 0.988 & 0.600 & 3.39\\
& \textbf{PersuaRL (Phi 3 mini)}    & \textbf{0.374} & \textbf{0.261} & \textbf{0.762} & \textbf{0.990} & \textbf{0.611} & \textbf{3.86}\\
\cmidrule(lr){2-8}
& Mistral 24B  Instruct (Single)    & 0.043 & 0.094 & 0.772 & 0.898 & 0.195 & 3.02\\
& Mistral 24B Instruct (SFT)       & 0.324 & 0.226 & 0.815 & 0.990 & 0.574 & 3.65\\
& \textbf{PersuaRL (Mistral 24B)}  & \textbf{0.355} & \textbf{0.241} & \textbf{0.873} & \textbf{0.992} & \textbf{0.596} &\textbf{4.12}\\
\midrule
\multirow{7}{*}{\parbox[c]{1.4cm}{\centering\textbf{DEAL}\\\footnotesize\cite{priya2024trip}}}
& Llama 3.2 3B instruct (Single)    & 0.049 & 0.085 & 0.519 & 0.937 & 0.195 & 2.41\\
& Llama 3.2 3B instruct (SFT)       & 0.080 & \textbf{0.104} & \textbf{0.552} & 0.986 & 0.267 & 2.64\\
& \textbf{PersuaRL (Llama 3.2 3B)}  & \textbf{0.087} & 0.101 & 0.536 & \textbf{0.987} & \textbf{0.278} & \textbf{2.79}\\
\cmidrule(lr){2-8}
& Phi 3 mini 128k (Single)          & 0.086 & 0.106 & 0.552 & 0.973 & 0.265 & 2.37\\
& Phi 3 mini 128k (SFT)             & 0.089 & 0.110 & 0.558 & 0.984 & 0.273 & 2.56\\
& \textbf{PersuaRL (Phi 3 mini)}    & \textbf{0.094} & \textbf{0.121} & \textbf{0.568} & \textbf{0.985} & \textbf{0.281} &\textbf{2.72}\\
\bottomrule
\end{tabular}%
}
\end{minipage}
\caption{Automatic evaluation results for \textbf{\textit{InsureDial}} and DEAL datasets. Bold values indicate the best performance within each model group. Results are statistically significant at 5\% significance level based on t-test. LLM-as-a-Judge scores are omitted for GPT models to avoid evaluation bias, as the underlying judge model is also GPT-based.}
\label{tab:model-comparison-sidebyside}
\end{table*}

\subsection{Baselines Setup}
We evaluate both closed-source and open-weight baselines in a single-shot setting, alongside a supervised fine-tuning (SFT) baseline. In the single-shot setup, models generate responses directly from the input without tool usage or intermediate reasoning. 
The comparison includes both open weight and proprietary models.
To assess portability, we instantiate \textbf{\textit{PersuaRL}} on open-weight backbones: \textsc{Llama-3.2-3B-Instruct}~\cite{meta2024llama3_2_3b_instruct}, \textsc{Phi-3-mini}~\cite{abdin2024phi3}, \textsc{Qwen-2.5-3B-Instruct}~\cite{qwen2}, and \textsc{Mistral-24B-Instruct}~\cite{mistral_small_3_2025}, evaluating under three regimes: (i) single-shot, (ii) SFT, and (iii)  \textbf{\textit{PersuaRL}}.
{\textit{Additional details regarding the baselines are mentioned in Appendix \ref{sec:Exp_detail}.}}

\subsection{Evaluation Metrics}

We conduct both automatic and human evaluations to assess the response quality. Automatic metrics include \textsc{ROUGE-1} (R1)~\cite{lin2004rouge}, \textsc{BLEU-2} (B2)~\cite{papineni2002bleu}, \textsc{METEOR} (MT)~\cite{banerjee2005meteor}, \textsc{BERT-F1} (BF1)~\cite{zhang2019bertscore}, \textsc{Distinct-2} (D2)~\cite{li2015diversity}, and \textsc{LLM-as-a-judge} (LLM-J) \cite{openai2025gpt5}. Human evaluation is performed across five dimensions{\footnote{Human evaluation was conducted using a 5-point scale, where 1 indicates the lowest and 5 the highest performance.}: Fluency (F), Engagingness (E), Persuasive Effectiveness (PE), Strategy Appropriateness (SA), and Resistance Handling (RH). \textit{Full details are provided in the Appendix under the section \ref{sec:Exp_detail}.}



\section{Results and Findings}
Table \ref{tab:model-comparison-sidebyside} presents the automatic evaluation results for our framework, \textbf{\textit{PersuaRL}}, alongside multiple baselines and serves as the basis for addressing the research questions outlined below.

\subsection{Research Questions}
\textbf{R1) How good is \textbf{\textit{PersuaRL}} compared to the baselines?}  
\textbf{\textit{PersuaRL}} \textbf{(Mistral)} consistently outperforms all baselines on the \textbf{\textit{InsureDial}} dataset as shown in Table \ref{tab:model-comparison-sidebyside}. Specifically, small backbone like \textbf{Llama~3.2--3B~\textit{PersuaRL}} achieves the result of \textsc{BF1}~$\approx$~0.771, \textsc{B2}~$\approx$~0.398, and \textsc{R1}~$\approx$~0.631, outperforming much larger 14--70B vanilla baseline models. For example, \textbf{\textit{PersuaRL}} (Phi-3 Mini) exceeds Phi-3 Medium 14B by roughly 16\% and Qwen-3 32B by over 29\% in BERT-F1. Although SFT helps models adapt to the insurance domain, its improvements largely saturate at surface-level generation. In contrast, \textbf{\textit{PersuaRL }}consistently pushes the models beyond this plateau. 



\textbf{R2) Can reward-guided reinforcement learning for expert selection outperform heuristic, prompt-based routing in persuasive dialogue systems?} \textbf{\textit{PersuaRL}} surpasses prompt-based routing by framing expert selection as a learnable action optimized through reward-driven credit assignment. While prompt-based routers rely on fixed heuristics without feedback, \textbf{\textit{PersuaRL}} uses GRPO to iteratively learn which expert combinations maximize persuasive effectiveness, enabling adaptive and context-sensitive expert coordination.

\textbf{R3) How robust and transferable is \textbf{\textit{PersuaRL}}? } To rigorously assess the robustness and cross-domain transferability of \textbf{\textit{PersuaRL}}, we conducted evaluations on the out-of-domain \textsc{DEAL} dataset, as shown in Table~\ref{tab:model-comparison-sidebyside}. Across all backbone configurations (\textit{Single} $\rightarrow$ \textit{SFT} $\rightarrow$ \textit{PersuaRL}), we observe consistent and monotonic improvements, reflected in higher \textsc{BF1} and \textsc{R1} scores.
Beyond aggregate performance, we further analyze the nature of transfer across domains. Despite differences in domain semantics (insurance vs. travel), annotation taxonomies, and user objectives, \textbf{\textit{PersuaRL}} maintains strong performance, suggesting that the learned selector policies capture domain-agnostic persuasion patterns such as intent adaptation, and strategic framing rather than overfitting to dataset-specific heuristics. To better understand transfer behavior, we conducted a detailed error analysis (see Table~\ref{tab:error_deal}). These findings indicate that \textbf{\textit{PersuaRL}} learns a combination of domain-agnostic coordination strategies (reward models) and domain-sensitive adaptations(generator), supporting its robustness and applicability across diverse dialogue settings.

\begin{figure*}[t]
\centering
\includegraphics[width=1\textwidth]{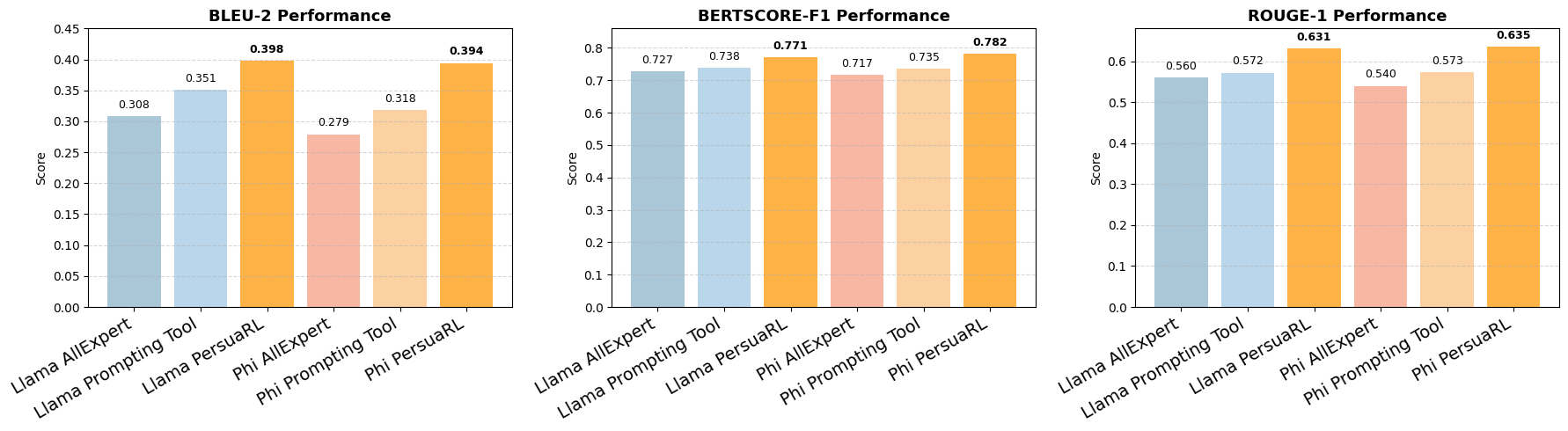} 
\caption{Ablation Study of Llama 3B and Phi 3B variants under (a) All Tools (no selector), (b) Tools Selected via Prompting, (c) Tool Selection via Reward-Driven \textbf{\textit{PersuaRL}}.}
\label{fig2}
\end{figure*}

\section{Ablation Studies}
\textbf{Ablation study w.r.t Experts: }
The full \textbf{\textit{PersuaRL}} model achieves the best overall performance, indicating that jointly leveraging all experts yields more persuasive responses. Removing any expert leads to consistent degradation, confirming that each module contributes meaningfully to the system. Results are reported in Table \ref{tab:ablation expert} using automatic metrics.
Among the experts, the Engagement and Intent modules have the most pronounced impact. Excluding either results in notable drops in semantic alignment and relevance. Although the keyterm expert and sentiment expert's individual impact is comparatively smaller, its removal still leads to a measurable decline in the overall quality.

\begin{table}[h!]
\centering

\resizebox{\linewidth}{!}{
\begin{tabular}{l c c c c c}
\hline
\textbf{Models} & \textbf{B-2 $\uparrow$} &\textbf{MT $\uparrow$} & \textbf{BF1 $\uparrow$} & \textbf{D-2 $\uparrow$} & \textbf{R1 $\uparrow$} \\
\hline
\multicolumn{6}{c}{\textbf{InsureDial}} \\
\hline
\textbf{\textit{PersuaRL}}        & \textbf{0.375}      &  \textbf{0.250}     &  \textbf{0.760}     &  \textbf{0.991}     &   \textbf{0.609}\\
\textbf{\textit{PersuaRL}} - Engagement Expert       & 0.284 & 0.202 & 0.704 & 0.983 & 0.539 \\
\textbf{\textit{PersuaRL}} - Intent Expert        & 0.293 & 0.216 & 0.713 & 0.983 & 0.558 \\
\textbf{\textit{PersuaRL}} - Keyterm Expert           & 0.302 & 0.223 & 0.721 & 0.984 & 0.566 \\
\textbf{\textit{PersuaRL}} - Sentiment Expert           & 0.324 & 0.231 & 0.736 & 0.990 & 0.583 \\
\hline

\end{tabular}}
\caption{Ablation study of expert modules for \textbf{\textit{PersuaRL}} on the Qwen 2.5 3B model.}
\label{tab:ablation expert}
\end{table}

\textbf{All Tools vs \textbf{\textit{PersuaRL} :}} We evaluate \textbf{\textit{PersuaRL}} against the \textbf{AllExpert} baseline, which uniformly activates all experts, across both \textsc{Phi} and \textsc{Llama-3B} LLMs. This ablation shows the \textit{impact of the selector} in our framework. \textbf{\textit{PersuaRL}} consistently outperforms AllExpert across all metrics, as shown in Figure \ref{fig2}, demonstrating strong generalization and better alignment with task preferences. Notably, it shows substantial gains in generation quality, particularly in semantic relevance and lexical overlap, with improvements most pronounced in R1 and BF1. 

\textbf{Prompting Tool Selection Vs \textbf{\textit{PersuaRL}} :} \textbf{\textit{PersuaRL}} outperforms prompt-based routing across both \textsc{Phi} and \textsc{Llama}-3B backbones, achieving higher fluency, relevance, and overall generation quality through preference-aligned expert coordination. These gains are consistent and do not compromise output diversity, demonstrating the effectiveness of reinforcement learning over generic routing strategies.



\section{Human Evaluation}
A team of annotators with domain expertise in insurance dialogue evaluation conducted a human assessment on 30\% of the randomly sampled \textit{\textbf{InsureDial}} test set. We evaluated three variants for each LLM (Llama~3.2~3B, Qwen~2.5~3B, and Mistral 24B): \textit{Single-shot}, \textit{SFT-finetuned}, and our proposed \textbf{\textit{PersuaRL}} framework. As shown in Table \ref{tab:Human-Evaluation}, \textbf{\textit{PersuaRL}} achieves the highest scores across all five dimensions for all the backbones, indicating consistent improvements in naturalness, engagement, and persuasion. 

\begin{table}[t]
\centering
\normalsize
\resizebox{\columnwidth}{!}{%
\begin{tabular}{lccccc}
\toprule
\textbf{Models} & \textbf{F} $\uparrow$ & \textbf{E} $\uparrow$ & \textbf{PE} $\uparrow$ & \textbf{SA} $\uparrow$ & \textbf{RH} $\uparrow$ \\
\toprule
\multicolumn{6}{c}{\textbf{InsureDial}} \\
\toprule
Llama 3.2 3B Instruct (Single)  & 2.47 & 2.31 & 2.13 & 2.39 & 2.94 \\
Llama 3.2 3B Instruct (SFT)     & 3.11 & 2.94 & 2.46 & 2.88 & 3.38 \\
\textbf{PersuaRL (Llama 3.2 3B)} & \textbf{4.12} & \textbf{4.51} & \textbf{4.36} & \textbf{4.29} & \textbf{4.46} \\
\toprule
Qwen 2.5 3B Instruct (Single)   & 2.69 & 2.45 & 2.29 & 2.68 & 3.10 \\
Qwen 2.5 3B Instruct (SFT)      & 3.19 & 2.98 & 2.86 & 3.10 & 3.61 \\
\textbf{PersuaRL (Qwen 2.5 3B)} & \textbf{3.94} & \textbf{4.22} & \textbf{4.23} & \textbf{4.06} & \textbf{4.32} \\
\toprule
Mistral 24B Instruct (Single)  & 2.98 & 2.81 & 2.53 & 2.74 & 3.27\\
Mistral 24B Instruct (SFT)     & 3.23 & 3.34 & 3.10 & 3.19 & 3.76 \\
\textbf{PersuaRL (Mistral 24B)} & \textbf{4.26} & \textbf{4.39} & \textbf{4.54} & \textbf{4.33} & \textbf{4.45}\\
\bottomrule
\end{tabular}%
}
\caption{Human evaluation result on \textbf{\textit{InsureDial}} datasets. The Single → SFT → PersuaRL trend
holds consistently at 3B–24B scale.}
\label{tab:Human-Evaluation}
\end{table}

\section{Qualitative Analysis}
Figure \ref{Qualitative_Analysis} and Table \ref{tab:insuredial_analysis} highlights qualitative differences between responses from single-shot, SFT, and \textbf{\textit{PersuaRL}}, demonstrating two key distinctions:
(1) \textbf{\textit{PersuaRL}} generates more empathetic and persuasive responses, as seen in: “I completely understand how important it is to feel safe and supported during your weekend rides,” followed by reassurances like “help is always just a call away” and “we’ve got your back wherever the road takes you”.  In contrast, SFT responses are more informative yet neutral merely listing features without user-centric framing.
(2) \textbf{\textit{PersuaRL}} also demonstrates greater fluency and structure, combining coverage details with contextual reassurance. For example, it fluidly mentions “third-party, own damage, and personal accident coverage but also 24/7 roadside assistance,” linking technical information with persuasive language. On the other hand, SFT and single shot give factual completeness but miss emotional framing. Additional qualitative examples generated by \textbf{\textit{PersuaRL}} are provided in Appendix Table \ref{tab:persuarl_phi_results} and Table \ref{tab:persuarl_qwen_results}.

\begin{figure}[t]
\centering
\includegraphics[width=0.5\textwidth]{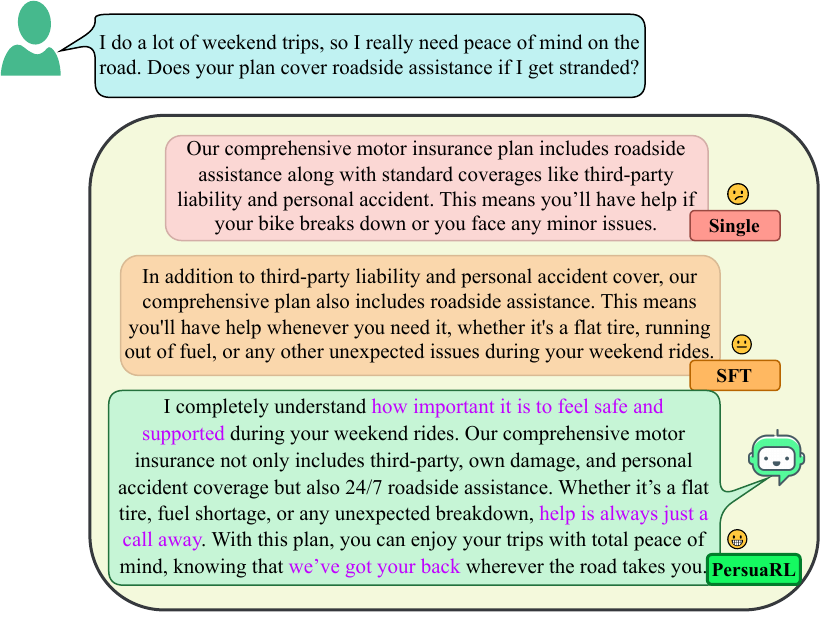} 
\caption{Qualitative analysis for an input utterance, response by single, SFT and our \textbf{\textit{PersuaRL.}}}
\label{Qualitative_Analysis}
\end{figure}

\section {Conclusion}
In this work, we proposed \textbf{\textit{PersuaRL}}, a reinforcement learning-based framework for persuasive dialogue generation in the insurance domain. Through modular expert design and reward-guided selection, \textbf{\textit{PersuaRL}} enables context-aware generation across LLM backbones. We also introduced \textbf{\textit{InsureDial}}, a high-quality annotated dataset for insurance persuasion. \textbf{\textit{PersuaRL}} consistently outperforms strong baselines across single-shot, SFT, and RL settings, including larger models, while producing fluent, persuasive, and emotionally aligned responses, highlighting its effectiveness for modular, goal-driven LLMs.


\section{Limitations}
The limitations of this work are stated in the below points : \par
\begin{itemize}[leftmargin=*]
    \item \textbf{\textit{InsureDial}} is constructed using a semi-automated pipeline where GPT-4o generates dialogues that are subsequently filtered by humans; this may introduce synthetic artifacts and may not fully capture real user behavior.
    \item The expert-selection policy operates over a binary mask, causing the action space to grow exponentially as the number of experts increases, which can hinder scalability despite the stabilizing effect of GRPO.
    \item Invoking multiple expert modules per dialogue turn increases inference latency, computational cost, and system complexity compared to single-model baselines, potentially impacting real-world deployment.
    \item All evaluation is offline and conditioned on gold dialogue history, so the model's own responses never shape subsequent turns. We, therefore, do not assess interactive rollouts or real-user outcomes; live user studies remain future work.
    \item While we evaluate \textbf{\textit{PersuaRL}} on smaller open-weight backbones (3B--24B), applying the full framework with large-scale LLMs as both the selector and generator was not feasible due to computational resource constraints, as this demands substantial GPU memory and compute. We note this as a promising direction for future work.
\end{itemize}

\section{Ethics Section}
This work emphasizes ethical persuasion by treating it as decision support rather than decision enforcement, ensuring that user autonomy is preserved and that responses align with user intent, sentiment, and context without applying coercive or repetitive pressure. Given high-stakes domains, where inaccuracies can lead to harm, particular care is taken to ensure factual accuracy and prevent misleading information\cite{sahoo2024comprehensive,ghosh2025clinic}. To mitigate such risks, \textbf{\textit{InsureDial}} was developed using a human-in-the-loop process, with domain experts validating seed dialogues and annotators verifying model-generated conversations, and explicitly annotating strategies and intents to support context-aware persuasion across multi-turn interactions.

\section*{Acknowledgement}
This research was conducted as part of the project ``Conversational Agents with Negotiation and Influencing Ability'', sponsored by Accenture Labs, Bangalore, India. The authors also acknowledge the National Supercomputing Mission (NSM) for providing computing resources on the PARAM Rudra supercomputer.


\bibliography{custom}

\appendix
\clearpage
\textbf{Frequently Asked Questions}
\label{sec:Freq_Ques}

\textbf{1. Is the strategy-consistency reward misaligned since persuasion strategies are annotated on agent turns, not user turns? }\par
\textbf{Response:} The strategy-consistency reward is used as a soft alignment signal rather than a hard supervisory target. While strategy annotations are available for agent turns, conditioning on the user utterance encourages the selector to anticipate the appropriate persuasive response given the user’s expressed needs. We acknowledge this approximation and view it as a practical surrogate that avoids requiring gold strategy labels at inference time.\par

\textbf{2. Are the reward equations underspecified, particularly with respect to class probabilities and weights?}\par
\textbf{Response:} The reward formulation uses class-conditional probabilities from pretrained classifiers; the omission of explicit class indices in the main text was for brevity. All rewards are computed using the predicted probability of the target class, and weights are normalized to sum to one. \par

\textbf{3. Does the contextual similarity reward encourage copying from the dialogue context or user utterance? } \par
\textbf{Response:} To mitigate copying, the contextual reward is complemented by a non-repetitiveness penalty that discourages lexical overlap with prior agent responses. Empirically, we observe improvements in diversity metrics (DISTINCT-2) alongside relevance gains, indicating that the model does not simply copy input text.\par

\textbf{4. Is the selector architecture and GRPO training pipeline sufficiently specified for reproducibility?}\par
\textbf{Response:} The selector is implemented as a lightweight transformer-based policy operating over a binary expert mask. GRPO is used with fixed group size, clipping range, and KL regularization. While the main paper focuses on conceptual clarity, all architectural details, hyperparameters, and training phases are provided in the appendix.\par

\textbf{5. Why are some closely related persuasion and routing baselines not included? }\par
\textbf{Response :} Our primary goal is to compare against strong single-shot, SFT, and heuristic-routing baselines under identical backbones. Many intent-to-strategy and persona-aware models rely on additional supervision or assumptions not directly comparable to our expert-selection setting. \par

\textbf{6. Does reinforcement learning improve persuasion or just optimize reward models?} \par
\textbf{Response :} \textbf{\textit{PersuaRL}} applies reinforcement learning only to \textit{expert-selection} actions. The generator is held fixed within each GRPO step and is never updated on the reward directly; it is updated only by supervised fine-tuning on the ground-truth response given the highest-reward expert selection. This limits the policy's ability to exploit reward artifacts. Despite relying on rewards, the learned selection policy consistently improves independent human evaluated metrics, indicating that reward-optimized expert generalizes beyond the reward models.

\textbf{7. How does \textbf{\textit{PersuaRL}} avoid reward circularity during training?} \par
\textbf{Response:} Reward circularity would occur if the model were trained to generate responses that directly satisfy the same models used to evaluate them. \textbf{\textit{PersuaRL}} avoids this by keeping the reward models frozen throughout and applying reinforcement learning only to \textit{expert-selection decisions}, a discrete action space of $2^n$ masks that cannot shape response text directly. The generator is never optimized against the reward: its parameters are updated only by supervised fine-tuning toward ground-truth responses, and the reward's sole influence on this step is the choice of which expert-augmented input is paired with the gold response. Reward therefore reaches the selector only indirectly, through the effect of its selection on the generator's output, and cannot be exploited by altering the generator's output distribution.

\textbf{8. How robust is the system to missing or noisy experts?}\par
\textbf{Response:} Robustness is partially evaluated through ablations that remove individual experts, showing graceful degradation rather than collapse. While we do not explicitly simulate noisy expert outputs, this remains an important extension for future study.\par
\textbf{9. Does \textit{PersuaRL} introduce excessive inference overhead?} \par
\textbf{Response:}  \textbf{\textit{PersuaRL}} trades additional computation for improved persuasion quality. In practice, only a small subset of experts is selected per turn, and all experts are lightweight models. On average, PersuaRL is taking 1.4 times more inference time than the SFT baselines. We report this as a conscious design trade-off rather than a limitation of feasibility.\par

\textbf{10. How should the results be interpreted beyond the insurance domain?}\par

\textbf{Response :} \textbf{\textit{InsureDial}} serves as a controlled, high-stakes testbed for persuasion. While absolute performance may not transfer directly, the formulation of expert selection as a learnable policy is domain-agnostic and can be applied to other persuasive dialogue settings, as shown by our robustness study.

\textbf{11. How does the \textit{PersuaRL} framework transfer to out-of-domain datasets, and what is the motivation for selecting DEAL as the out-of-domain benchmark?
}

\textbf{Response :}The core contribution of \textbf{\textit{PersuaRL}} is the selector policy that formulates expert coordination as a learnable decision-making problem. This formulation is domain-agnostic in the sense that it treats expert coordination as a reinforcement learning problem operating over expert signals and reward feedback, independent of insurance-specific semantics. However, the learned policy is influenced by the reward signals and expert training specific to the motor-insurance setting. We selected DEAL as a baseline because it is a multi-turn negotiation benchmark involving persuasion dynamics such as resistance handling, objective trade-offs, and strategic adaptation. Our goal was not to claim universal persuasion generalization, but to evaluate whether the expert-coordination mechanism continues to provide benefit under domain shift. 

\clearpage

\section{Appendix}
\label{sec:appendix}

This appendix provides supplementary material, including detailed descriptions of dataset construction (Section~\ref{dataset_cons}), experimental settings and implementation details (Section~\ref{sec:Exp_detail}), additional experimental results (Section~\ref{sec:add_exp}), and the full set of prompts used in our experiments (Section~\ref{sec:Prompts}).

\section{Dataset Construction}
\label{dataset_cons}
\subsection{Annotation guidelines}

A critical part of our pipeline was comprehensive annotator training and clear instructions. Before beginning annotation, annotators were explicitly trained using sample dialogues illustrating each persuasion strategy, user intent, sentiment, and key term category. They were provided with detailed annotation guidelines, including definitions, decision rules, and examples for every label. This ensured a shared understanding and consistency across annotators from the outset. The dataset was annotated across four key dimensions: (i) Engagement Strategy (Logical, Credibility, Emotional, Personal, Persona, Default), (ii) User Intent (Request Quote, Ask Coverage Details, Express Concern, Request Additional Info, Confirm Interest, Ask Price/Premium), (iii) Sentiment (Positive, Neutral, Negative), and (iv) Key Domain Terms (e.g., “No Claim Bonus,” “Roadside Assistance,” “Depreciation”).
\subsection{Dataset Annotation}
\label{sec:Dataset_Annotation}
The \textbf{\textit{InsureDial}} dataset was annotated through a carefully staged, human-in-the-loop process involving five trained human annotators. We began by manually annotating 75 dialogues, where each utterance was labelled across four dimensions: engagement strategy, user intent, sentiment, and key domain terms. These 75 dialogues served as gold-standard references for subsequent semi-automated annotation. An initial labeling pass was performed using Gemini-2.0-Flash \cite{google_gemini_2_0_flash} for efficiency, and all labels were subsequently verified and refined by human annotators to ensure correctness and contextual alignment. To achieve this, we designed few-shot prompts using examples drawn from the initial 75 human-annotated dialogues, enabling Gemini to accurately assign labels. As a quality safeguard, we first conducted a pilot annotation of 30 dialogues with Gemini, which were fully reviewed by human annotators. Only after the annotators confirmed the correctness and reliability of these annotations did we proceed to annotate the rest of the dataset, maintaining human verification in the loop to ensure high-quality and contextually accurate labels. A representative example from the \textbf{\textit{InsureDial}} dataset is shown in Table \ref{tab:insuredial_samples}.

\begin{table*}[h!]
\centering
\small 
\renewcommand{\arraystretch}{1.5} 

\begin{tabularx}{\textwidth}{|>{\hsize=1.4\hsize}X|>{\hsize=0.6\hsize}X|} 
\hline
\textbf{User Utterances} & \textbf{Keyterms} \\ \hline

\textit{Hi, I'm looking for a motor insurance policy for my bike. It's a 2022 Royal Enfield Classic 350.} & 
Motor insurance, 2022 Royal Enfield Classic 350 \\ \hline

\textit{That sounds good. What about roadside assistance? I've heard Teslas can sometimes have issues.} & 
Roadside assistance \\ \hline

\textit{Okay, 'user-friendly' sounds good. What about discounts? As a student, every bit helps.} & 
User-friendly, Discounts \\ \hline

\textit{Hi, I need to get a new motor insurance policy for my 2020 Hyundai Elantra.} & 
Motor insurance, 2020 Hyundai Elantra \\ \hline

\textit{Personal Accident Cover sounds important. What else should I consider for a bike like mine?} & 
Personal Accident Cover \\ \hline

\textit{Zero Depreciation and Roadside Assistance... what exactly do those do?} & 
Zero Depreciation, Roadside Assistance \\ \hline

\textit{Paperless claims sounds ideal. I'm not a fan of excessive forms. Can you send me the details to review?} & 
Paperless claims \\ \hline

\end{tabularx}
\caption{Critical keyterms pertaining to the motor insurance domain.}
\label{tab:keyterm}
\end{table*}

\begin{table*}[h!]
\centering
\small
\renewcommand{\arraystretch}{1.5} 

\begin{tabularx}{\textwidth}{|>{\hsize=0.4\hsize}X|>{\hsize=1.6\hsize}X|} \hline
\textbf{Engagement Expert} & \textbf{Agent Utterances} \\ \hline

Logical Appeal & \textit{Welcome! I'm glad to help you with motor insurance for your new Tesla Model 3. Our policies offer comprehensive coverage for electric vehicles, including battery protection. Let's discuss the options that best suit your needs.} \\ \hline

Credibility Appeal & \textit{Tata AIG has built a strong reputation in the insurance industry for over two decades. Our commitment to innovation, customer-centric approach, and robust risk management practices make us a reliable choice for motor insurance. We leverage modern technology to streamline processes, ensuring a seamless and hassle-free experience for our customers. With Tata AIG, you can trust that your Tesla Model 3 and your family are in good hands.} \\ \hline

Emotional Appeal & \textit{I completely understand your concerns about keeping your family safe on the road. Our comprehensive coverage options for your 2010 Honda Accord include third-party liability, own damage, and add-on covers to ensure maximum protection for you and your loved ones. Let's discuss the best plan that suits your needs and provides peace of mind while driving.} \\ \hline

Persona-Based Appeal & \textit{I'm delighted to help you with your insurance needs for your new 2024 Tesla Model 3. To provide you with an accurate quote, I'll need a few details about your driving history and preferences. Once I have that information, I can offer you a tailored policy that suits your needs and budget. Let's get started!} \\ \hline

Personal Appeal & \textit{Reliance General Insurance offers competitive rates and a range of coverage options for your 2018 Maruti Suzuki Baleno. As a valued customer, you may be eligible for loyalty discounts. Let's explore the benefits and coverage details of Reliance General Insurance to see if it's the right fit for you and your budget.} \\ \hline

Default & \textit{I'm glad you're considering our insurance for your 2010 Maruti Suzuki Swift. Regarding the claim process, we strive to make it as simple and straightforward as possible. You can file a claim through our website or by calling our customer service line. Our team is always ready to assist you, and we'll guide you through the process step by step. If you have any concerns or questions, please don't hesitate to ask. We're here to help.} \\ \hline

\end{tabularx}
\caption{Example agent utterances for different engagement expert strategies in the insurance domain.}
\label{tab:engagement_examples}
\end{table*}

\begin{table*}[h!]
\centering
\small 
\renewcommand{\arraystretch}{1.5} 

\begin{tabularx}{\textwidth}{|>{\hsize=0.3\hsize}X|>{\hsize=1.7\hsize}X|}
\hline
\multicolumn{2}{|c|}{\textbf{InsureDial}} \\ \hline

\textbf{Context} & 
\textbf{User}: \textit{Hello, I need to insure my 2022 Tesla Model 3. It's a new car, and I want the best coverage possible.} \newline
\textbf{Agent}: \textit{I'm delighted to help you with your insurance needs for your new 2022 Tesla Model 3. We offer comprehensive coverage options tailored to electric vehicles, ensuring your car is protected in all key areas. Let's discuss the best plan for you to provide the peace of mind you deserve.} \newline
\textbf{User}: \textit{I'm a business owner, 45 years old, and I primarily drive the Tesla on weekends with my family. Safety for them is paramount, and I don't mind paying for premium protection.} \newline
\textbf{Agent}: \textit{Given your role as a business owner and the importance you place on safety for your family, I'm glad to inform you that our comprehensive EV coverage includes features designed to prioritize your loved ones' well-being. We'll work together to find a plan that offers premium protection for your 2022 Tesla Model 3, ensuring peace of mind for both your personal and professional life. Let's discuss the details further to find the perfect fit for your needs. \newline
\textbf{User}: That sounds promising. What makes it ``premium'' protection? I want comprehensive coverage, of course.} \\ \hline

\textbf{Gold Response} & \textit{Our premium protection goes beyond standard coverage, offering enhanced features such as extended warranty, roadside assistance, and personal accident coverage. This comprehensive coverage ensures that you and your family are well-protected in various scenarios, providing the peace of mind you seek for your 2022 Tesla Model 3. Let's delve deeper into the specifics of our premium protection plan to find the best fit for your needs.} \\ \hline

\textbf{Single} & Premium protection just means your Tesla will be covered for accidents and other situations. You will have comprehensive coverage that should keep your car insured. \\ \hline

\textbf{SFT} & \textit{Our premium protection plan offers full accident coverage, roadside assistance, and protection for both the vehicle and passengers. It’s designed to provide complete comprehensive coverage for Tesla owners like you. You’ll have strong protection whether driving for personal use or with family on weekends.} \\ \hline

\textbf{\textit{PersuaRL}} &\textit{Our premium protection is built to give you total confidence every time you take your Tesla out. Beyond standard coverage, it includes enhanced passenger safety, extended vehicle protection, roadside assistance, and personal accident benefits. Whether it’s a weekend trip with your family or a daily errand, you’ll enjoy complete peace of mind knowing your 2022 Tesla Model 3 and your loved ones are fully shielded. Let’s explore the details together to ensure you get the absolute best protection for your lifestyle.} \\ \hline

\end{tabularx}
\caption{Qualitative Analysis}
\label{tab:insuredial_analysis}
\end{table*}


\section{Experiment Details}
\label{sec:Exp_detail}
\textbf{Baselines Details.} In addition to implementing \textbf{\textit{PersuaRL}} on smaller language models, we also incorporated large scale baseline models, including both open source and proprietary systems, as shown below.

\begin{enumerate}
 \item \textbf{Closed-Source Models:} GPT 5 \cite{openai2025gpt5}, GPT-4.1 Mini \cite{openai2025gpt4_1_mini}

 \item \textbf{Open-Weight Models:} DeepSeek-R1-Distill-LLaMA 70B \cite{deepseekai2025deepseekr1incentivizingreasoningcapability}, LLaMA-3.3-70B-Instruct \cite{meta2024llama3_3_70b}, Qwen-3-32B~\cite{qwen3technicalreport}, 
 Mistral-24B-Instruct \cite{mistral_small_3_2025}, Phi-3-Medium-14B \cite{abdin2024phi3}, LLaMA-3.1-8B-Instruct~\cite{meta2024llama3_1_8b_instruct}, Qwen-2.5-7B-Instruct~\cite{qwen2}

\end{enumerate}

We systematically compare single-shot generation, SFT, and PersuaRL across small open-weight models. Larger models (32B–70B) serve as high-capacity references, while our approach highlights \textbf{\textit{PersuaRL}}’s ability to achieve their performance even with compact models.

\subsection{Inference Regime}
Our experiments evaluate models under three distinct regimes, single-shot, SFT, and \textbf{\textit{PersuaRL}}, to provide a full picture of how different approaches impact persuasive dialogue generation.

\subsubsection{Single-Shot Generation.}
In the single-shot setting, the model is tasked with generating the agent’s next response based solely on the full dialogue context, which includes the entire conversation history and the current user utterance, without any task-specific fine-tuning. The model is provided with a simple instruction prompt alongside the conversation, and it produces a response in a single forward pass. In this regime, the model does not leverage intermediate reasoning, expert modules, or any reward-based selection, it simply relies on the general knowledge and capabilities acquired during pretraining. This setup establishes a baseline to assess how well large language models can generate contextually relevant and potentially persuasive responses without any explicit adaptation to the insurance dialogue domain.

\subsubsection{Supervised Fine-Tuning (SFT).}
SFT involves fine-tuning the base open-weight models on the \textbf{\textit{InsureDial}} dataset using a next-token prediction objective. This approach helps the model internalize domain-specific dialogue patterns and persuasion strategies.
Given a dialogue context $x = (u_1, a_1, \dots, u_t)$ and the target agent response $y = (y_1, y_2, \dots, y_T)$, the model is trained to maximize the likelihood of the next token:
\[
\mathcal{L} = -\sum_{t=1}^{T} \log P_\theta \left( y_t \mid x, y_{<t} \right).
\]
where $y_{<t}$ represents all previously generated tokens in the agent's response. We use the same simple conversation prompt as in the single-shot setting, but now the model has been fine-tuned to better predict the next agent utterance based on similar examples in the training set. SFT significantly improves response fluency and alignment with domain-specific persuasive strategies, but remains deterministic and does not dynamically adapt strategies at inference time beyond what it memorized during training.

\subsection{Expert Module}
\label{sec:Expert_Module}
The Expert Module in \textbf{\textit{PersuaRL}} leverages four specialized transformer-based models, Engagement Expert, Intent Expert, Keyterm Expert, and Sentiment Expert, to generate persuasive and context-aware responses. Each expert focuses on a distinct aspect of dialogue understanding and contributes unique insights to the final output.

\begin{table*}[h!]
\centering
\renewcommand{\arraystretch}{1.5} 

\begin{tabularx}{\textwidth}{|>{\hsize=0.6\hsize}X|>{\hsize=1.4\hsize}X|}
\hline
\textbf{Intent} & \textbf{User Utterances} \\ \hline

Request\_Insurance\_Quote & \textit{Hi, I'm looking to get insurance for my bike. It's a 2022 Royal Enfield Interceptor 650.} \\ \hline

Ask\_Coverage\_Details & \textit{What does the comprehensive plan cover exactly?} \\ \hline

Express\_Concern & \textit{Safety for my family is my top priority, but I also run a business, so I need to be mindful of the cost.} \\ \hline

Ask\_Additional\_Info & \textit{What about the company's reputation? I want to make sure they're reliable.} \\ \hline

Confirm\_Interest & \textit{That's a reasonable price. I'm happy to buy it online.} \\ \hline

Ask\_Price\_or\_Premium & \textit{What does the Tata AIG comprehensive policy cover specifically for electric vehicles?} \\ \hline

\end{tabularx}
\caption{Example user utterances for different intents in the insurance domain.}
\label{tab:intent_example}
\end{table*}

\begin{enumerate}
    \item \textbf{Engagement Expert} \\
    The Engagement Expert determines the most suitable \textit{persuasion strategy} for the current dialogue turn. 
    It classifies the context into six strategies, each defined as follows:
    \begin{enumerate}
        \item \textbf{Logical Appeal} – Employs factual reasoning, feature comparisons, or cost-benefit arguments to support recommendations.
        \item \textbf{Credibility Appeal} – Highlights trustworthiness, reliability, or reputation of the insurance provider.
        \item \textbf{Emotional Appeal} – Focuses on creating reassurance, security, or peace of mind for the user.
        \item \textbf{Personal Appeal} – Adapts responses to the user’s explicitly stated needs or concerns.
        \item \textbf{Persona Appeal} – Aligns policy recommendations with the user’s lifestyle, habits, or preferences.
        \item \textbf{Default (Neutral)} – Provides straightforward, informative responses without explicit persuasive framing.
    \end{enumerate}
    This expert is fine-tuned using negative log-likelihood (NLL) loss given by  $\quad L_{nll} = -\sum_{t=1}^{T} \log P_{\theta}(y_t \mid y_{<t}, x, \hat{k})$ respectively to balance accurate strategy classification and fluent generation. The examples of each strategy are shown in Table \ref{tab:engagement_examples}. 

    \item \textbf{Intent Expert} \\
    \label{sec:intent2}
    The Intent Expert identifies the \textit{user’s underlying goal} in each turn, which is crucial for generating contextually relevant and strategy-aligned responses. 
    Typical intents include \textit{Request Quote}, \textit{Ask Coverage Details}, \textit{Express Concern}, \textit{Request Additional Info}, \textit{Confirm Interest}, and \textit{Ask Price}. 
    Accurate intent detection allows the system to adapt its persuasive approach to the user’s decision-making stage. 
    This expert is trained with categorical cross-entropy and NLL loss, where $\quad L_{nll} = -\sum_{t=1}^{T} \log P_{\theta}(y_t \mid y_{<t}, x, \hat{i})$, to ensure robust classification and context-aware response generation. The examples of each strategy are shown in Table \ref{tab:intent_example}.

    \item \textbf{Keyterm Expert} \\
    \label{sec:keyterm2}
    The Keyterm Expert extracts \textit{critical domain-specific terms} from the user’s utterances, such as ``depreciation,'' ``roadside assistance,'' or ``personal accident coverage.'' 
    These key terms help the system highlight essential insurance concepts, making responses precise and informative. 
    The model is fine-tuned using next-token prediction on masked sequences labeled with key terms, producing structured outputs that are integrated by the Generator. The loss function is given by $L_{keyterm} = -\sum_{t=1}^{T} \log P(x_t \mid x_{<t})$. The examples of each strategy are shown in Table \ref{tab:keyterm}. 

    \item \textbf{Sentiment Expert} \\
    \label{sec:sentiment2}
    The Sentiment Expert detects the \textit{emotional tone} of the user’s message, classifying it as \textit{Positive}, \textit{Neutral}, or \textit{Negative}. 
    This enables sentiment-aware adaptation, such as providing empathy or reassurance when negative sentiment is detected. 
    By understanding emotion, the system can engage users more effectively and build trust during multi-turn dialogues. It enables sentiment-aware response generation, allowing the Engagement Expert to align its strategy with the user’s emotional state (e.g., emphasizing reassurance for negative sentiment). 
    This expert is trained with cross-entropy classification loss $L_{sentiment} = - \sum_{c \in \{pos, neg, neu\}} y_c \log \hat{y}_c$, to ensure robust sentiment recognition.
\end{enumerate}

\begin{figure*}[t]
\centering
\includegraphics[width=1\textwidth]{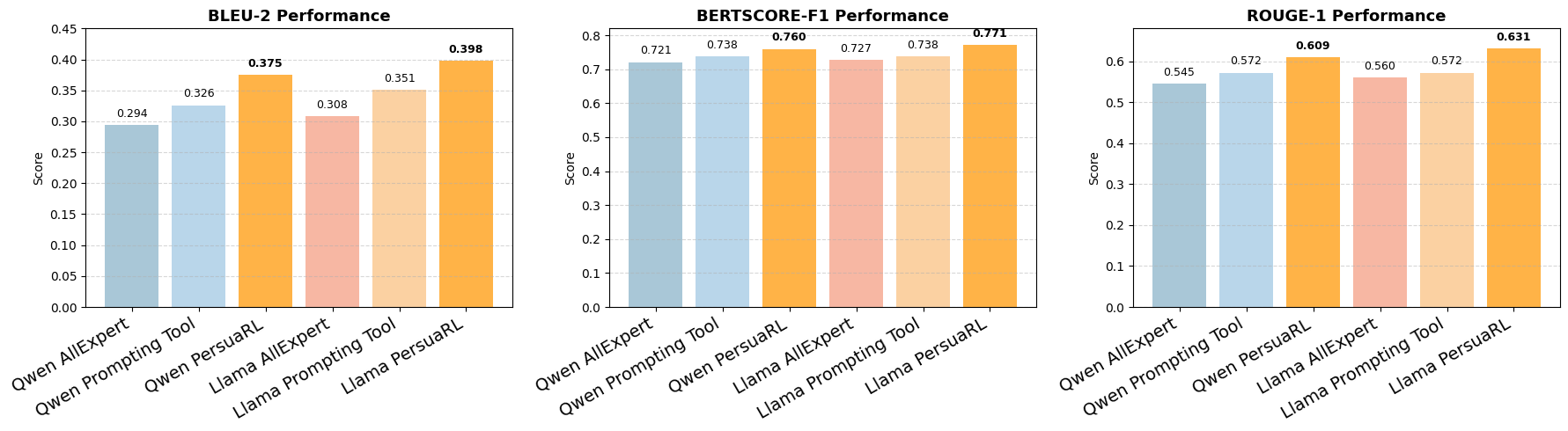} 
\caption{Ablation Study of Qwen 3B and Llama 3B variants under (a) All Tools (no selector), (b) Tools Selected via Prompting, (c) Tool Selection via Reward-Driven \textbf{\textit{PersuaRL}}.}
\label{Qwen_Llama}
\end{figure*}

\begin{figure*}[t]
\centering
\includegraphics[width=1\textwidth]{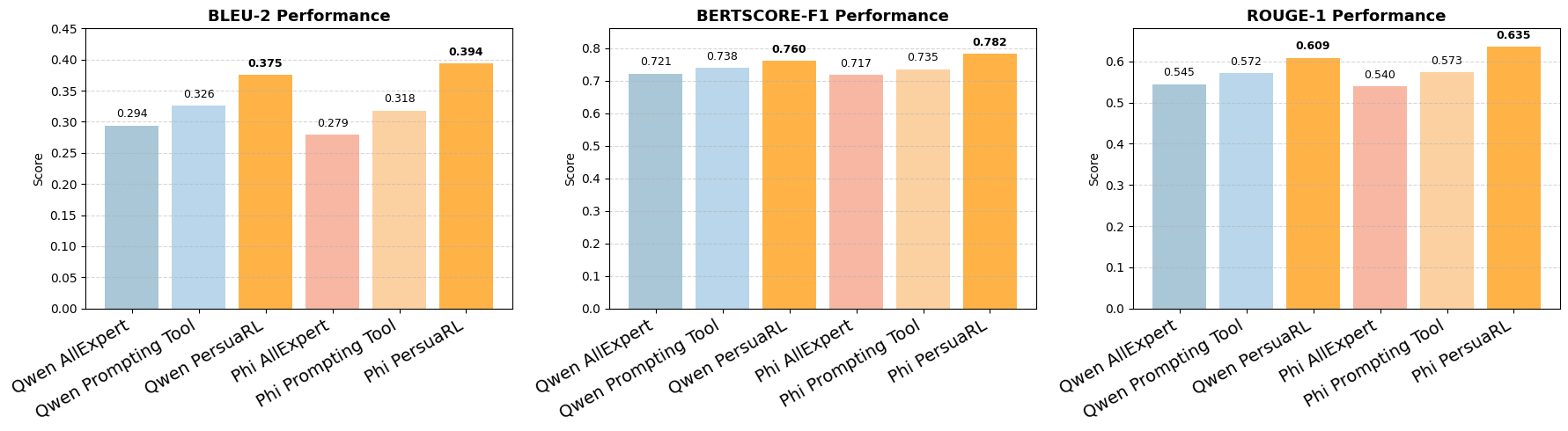} 
\caption{Ablation Study of Qwen 3B and Phi 3B variants under (a) All Tools (no selector), (b) Tools Selected via Prompting, (c) Tool Selection via Reward-Driven \textbf{\textit{PersuaRL}}.}
\label{Qwen_Phi}
\end{figure*}

\subsection{Classification Model.} 
We built the Engagement Strategy Consistency Reward (ESCR) and Intent Consistency Reward (ICR) by fine-tuning three pre-trained classifiers: BERT-Large \cite{devlin2019bert}, DistilBERT-base \cite{sanh2019distilbert}, and ModernBERT \cite{warner2024smarter}, on the \textit{\textbf{InsureDial}} dataset.
The performance of these classifiers is reported in Table \ref{tab:classifier_results}. It is evident that BERT outperforms the other models in terms of both Accuracy (Acc) and Macro F1 score.

\begin{table*}[t] 
\centering
\small
\begin{tabular}{lcccccc}
\toprule
\textbf{Classifier} 
& \multicolumn{2}{c}{\textbf{BERT-large}} 
& \multicolumn{2}{c}{\textbf{DistilBERT-base}} 
& \multicolumn{2}{c}{\textbf{ModernBERT}} \\
\cmidrule(lr){2-3} \cmidrule(lr){4-5} \cmidrule(lr){6-7}
& \textbf{Acc} & \textbf{F1} 
& \textbf{Acc} & \textbf{F1} 
& \textbf{Acc} & \textbf{F1} \\
\midrule
ESCR & 82.14 & 75.53 & 81.49 & 71.80 & 81.64 & 72.39 \\
ICR  & 84.91 & 74.49 & 81.52 & 72.34 & 80.78 & 70.26 \\

\bottomrule
\end{tabular}
\caption{Performance on the \textit{InsureDial} dataset (Accuracy and Macro F1).}
\label{tab:classifier_results}
\end{table*}

\subsection{Evaluation Details. }
We conducted a comprehensive evaluation of automatic and human metrics. For automatic metrics, scores are computed per agent turn and averaged across all turns in the test set. For human evaluation, Fluency, Engagingness, and Strategy Appropriateness are assessed at the turn level, while Persuasive Effectiveness and Resistance Handling are evaluated at the dialogue level.



\subsubsection{Human Evaluation Metrics}

Human evaluation is conducted across five dimensions to comprehensively assess the quality and effectiveness of generated responses. Fluency (F) measures the grammatical correctness, syntactic well-formedness, and overall readability of the response. Engagingness (E) evaluates how well the response sustains user interest by maintaining relevance and providing informative or contextually appropriate content. Persuasive Effectiveness (PE) measures how convincingly the response supports its intended argument. A highly persuasive response presents clear reasoning, compelling evidence and coherent argumentation that encourages the user to adopt the suggested viewpoint. Additionally, this metric evaluates whether the response appropriately incorporates relevant persuasion strategies or not.
Strategy Appropriateness (SA) evaluates whether the response uses the most suitable task-specific strategy for the given context. This includes selecting approaches that are goal-oriented, context-aware, and aligned with the expected dialogue style (e.g., logical, emotional, or credibility). Resistance Handling (RH) assesses the system’s ability to address user hesitations, objections, or counterarguments effectively. High performance in this metric involves acknowledging user concerns, providing contextually appropriate responses, and maintaining coherence without becoming confrontational or dismissive.

\subsection{Rewards and Penalties} 
\label{sec:penalties}

The additional details of the rewards are given below:

\textbf{Engagement Strategy Consistency Reward (R1):}
To ensure that the generated responses remain consistent with the user’s persuasion strategy, we fine-tune a BERT \cite{devlin2019bert} based persuasion strategy classifier (achieve 82.1\% accuracy on \textsc{InsureDial}) trained on six persuasion strategy labels $\mathcal{P} = \{0,1,2,3,4,5\}$; for each strategy $p \in \mathcal{P}$, we compute a persuasion prototype embedding as $\textit{PersuasionProto}_p = \frac{1}{N_p} \sum_{j=1}^{N_p} \textit{Embed}(u^p_j)$, where $N_p$ is the number of utterances labeled with strategy $p$. Given a generated response $r_T$, its compatibility score with strategy $p$ is $S_p(r_T) = \cos\big(\textit{Embed}(r_T), \textit{PersuasionProto}_p\big)$, and the persuasion strategy alignment reward is formulated as:

\begin{equation}
R_{\text{1}} = \sum_{p \in \mathcal{P}} P^{\text{pers}}_p(u_T) \cdot S_p(r_T) \,+\, \lambda \max_{p \in \mathcal{P}} S_p(r_T)
\end{equation}
where, $P^{\text{pers}}_p(u_T)$ is the BERT-predicted probability that the user utterance $u_T$ belongs to persuasion strategy $p$, and the first term encourages the generated response to align with the user’s persuasion strategy distribution.
We introduce $\lambda$ to balance alignment and confidence, where higher $\lambda$ emphasizes the most compatible class \footnote{$0.3< \lambda < 1$}.

\textbf{Intent Consistency Reward (R2):} To ensure that the generated responses remain consistent with the user’s intent, we fine-tune a BERT \cite{devlin2019bert} based intent classifier (achieve 84.9\% accuracy on \textsc{InsureDial}) trained on six user intent labels $\mathcal{I} = \{0,1,2,3,4,5\}$; for each intent $i \in \mathcal{I}$, we compute an intent prototype embedding as $\textit{IntentProto}_i = \frac{1}{N_i} \sum_{j=1}^{N_i} \textit{Embed}(u^i_j)$, where $N_i$ is the number of utterances labeled with intent $i$. Given a generated response $r_T$, its compatibility score with intent $i$ is defined as $S_i(r_T) = \cos\big(\textit{Embed}(r_T),\, \textit{IntentProto}_i\big)$, and the intent alignment reward is then formulated as:
\begin{equation}
R_{\text{2}} = \sum_{i \in \mathcal{I}} P^{\text{intent}}_i(u_T) \cdot S_i(r_T) \,+\, \lambda \max_{i \in \mathcal{I}} S_i(r_T)
\end{equation}
Here, $P^{\text{intent}}_i(u_T)$ represents the BERT-predicted probability of user utterance $u_T$ belonging to intent $i$. It encourages the generated response to align with the user’s intent distribution by weighting the compatibility score with the user’s intent probabilities.

\textbf{Contextual Appropriateness Reward (R3): }
To ensure that generated responses remain relevant to both the full dialogue context and the current user utterance, we define a reward function that leverages semantic similarity measured using BERT-F1 (BERTF1) \cite{zhang2019bertscore}. The reward encourages the model to produce responses that are contextually aligned while penalizing off-topic or irrelevant outputs. We give a higher weight to the current utterance (multiplied by 2) because relevance to the user’s last message is more crucial for perceived appropriateness.

\begin{equation}
R_3 = \min \Bigg( \frac{\text{BS}_{\text{F1}}(x_i, y_i) + 2 \cdot \text{BS}_{\text{F1}}(u_i, y_i)}{3}, \, 1 \Bigg)
\end{equation}
where, $x_i$ represents the entire dialogue context up to turn $i$, $u_i$ denotes the current user utterance at turn $i$, and $y_i$ is the model‑generated response for that turn.
We divide by 3 to normalize the weighted sum of \(\mathrm{BS}_{F1}\) values to the \([0,1]\) range, ensuring consistency with other rewards. The \(\min(\cdot,1)\) operation further limits the score to 1 to prevent outlier values from destabilizing reinforcement learning updates.

\textbf{Non-Repetitiveness Reward (R4): }
To encourage the model to generate diverse responses without repeating content from the previous turn, we define the non-repetitiveness reward $R_4$, which evaluates the lexical overlap between the current response $r_T$ and the previous response $r_{T-1}$ at consecutive dialogue turns $T$ and $T-1$:

\begin{equation}
R_4 = 1 - \frac{r_{T-1} \cap r_T}{r_{T-1} \cup r_T}
\end{equation}

\textbf{Judge Reward (R5): }
To capture high-level persuasive quality beyond surface-level lexical and classifier-based signals, we incorporate an LLM-as-a-judge reward using Prometheus-7B-v2.0 \cite{kim2024prometheus}. The judge model is prompted to act as a fair and objective evaluator, assessing each generated response with respect to persuasiveness, negotiation effectiveness, and user engagement in insurance sales. 
The reward graph is shown in Figure \ref{Reward Graph}.

The selector model decides which experts to consult before generating a response. While the main reward measures how good the final response is, additional penalties are introduced to shape the selector’s behavior. These terms act as soft constraints that encourage efficient reasoning and ensure balanced use of all available experts.

\begin{enumerate}
    \item \textbf{Complexity Penalty.} This penalty is motivated by the principle that more information is not always better. The goal is to discourage the selector from always choosing many experts when fewer would be sufficient. This penalty increases linearly as more experts are selected. Each additional expert slightly reduces the reward. This promotes simple and efficient expert selection. 

\[
\text{\textit{Complexity Penalty}} = \alpha \times N
\]

where:
\begin{itemize}
  \item $N$ is the number of experts selected in the current route.
  \item $\alpha = 0.025$ is a small constant penalty per expert.
\end{itemize}

\item \textbf{Route Repetition Penalty.}
In reinforcement learning, repeatedly selecting the same action can lead to policy collapse, where the model stops exploring alternative strategies. In this context, an action corresponds to selecting a particular route\footnote{A route refers to a specific combination of experts selected by the selector model for a given user utterance.}. The route repetition penalty discourages excessive reuse of the same route, encouraging the selector to explore different expert combinations. This allows the selector to discover different routing strategies that may perform better in diverse conversational contexts.

\[
\begin{aligned}
\textit{Repetition Penalty}
&= \min\!\Big( \beta \cdot \max(0, F - 1), \\
&\qquad\qquad P_{\max} \Big)
\end{aligned}
\]

\begin{itemize}
  \item $F$ is the ratio of route usage to ideal usage.
  \item $\beta = 0.2$ is the penalty scaling factor.
  \item $P_{\max} = 0.15$ is the maximum repetition penalty.
\end{itemize}

\item \textbf{Load Balance Penalty.}
Each expert is designed to capture a different aspect of the user’s intent or emotional state. If one expert is overused, the system becomes biased toward a narrow perspective and loses the benefits of specialization. The load balance penalty discourages long-term overuse of any single expert, ensuring that all experts remain meaningful contributors to the decision process.

\[
\textit{Load}_k =
\begin{cases}
\gamma \left( R_k - 1 \right)^2, & \text{if } R_k > 1, \\
0, & \text{otherwise}
\end{cases}
\]

\begin{itemize}
  \item $R_k$ is the number of times expert $k$ has been used so far, divided by the average number of times all other experts have been used.
  \item $\gamma = 0.4$ is a scaling constant that controls the strength of the load-balance penalty.
\end{itemize}

The load balance penalty is capped at 0.15.
\end{enumerate}

\section{Additional Experiments}
\label{sec:add_exp}
\subsection{Weight Optimization}

To determine the optimal combination of weights for the reward function, we performed experiments with different combinations of $\beta_1, \beta_2, \beta_3$, $\beta_4$ and $\beta_5$. These weights were validated using a 15\% hold-out \textbf{\textit{InsureDial}} dataset, and the combination that achieved the lowest perplexity score was selected for training \textbf{\textit{PersuaRL}}. Table \ref{tab:weight optimization} lists the weights considered for optimization using the \textbf{\textit{InsureDial}} dataset. The results in the table indicate that including all rewards yields a better perplexity score. Moreover, eliminating any single reward leads to a noticeable drop in the perplexity score (PPL), thereby emphasising the contribution of each reward to the overall model performance.

\begin{table}[h!]
\centering
\begin{tabular}{ccccccc}
\hline
\multicolumn{7}{c}{\textbf{WEIGHT OPTIMIZATION}} \\
\hline
$\beta_1$ & $\beta_2$ & $\beta_3$ & $\beta_4$ & $\beta_5$ & & PPL \\
\hline
0.1 & 0.15 & 0.35 & 0.3 & 0.1 & &  4.01654  \\
0.0 & 0.0 & 0.65 & 0.2 & 0.1 & &   4.0982  \\
0.0 & 0.0 & 0.45 & 0.5 & 0.05 & &   4.3182  \\
0.2 & 0.3 & 0.1 & 0.2 &0.2 & & 3.91726 \\
0.3 & 0.1 & 0.3 & 0.1 & 0.2 & & 3.91963 \\
0.1 & 0.25 & 0.15 & 0.35 & 0.15 & & 3.991 \\
0.25 & 0.2 & 0.2 & 0.2 & 0.15 & & 3.9183 \\
0.25 & 0.15 & 0.15 & 0.2 & 0.3 & & 3.9194 \\
\textbf{0.15} & \textbf{0.15} & \textbf{0.20} & \textbf{0.15} & \textbf{0.35}& & \textbf{3.91723} \\
\hline
\end{tabular}
\caption{Weight optimization using different reward weight combinations for Qwen 2.5 3B Instruct.}
\label{tab:weight optimization}
\end{table}


\subsection{Experimental Setup}
\label{sec:Experimental_Setup}
All experiments were conducted on an A100 80GB GPU, with each model training
for approximately 25--28 hours. Implementations were done in
PyTorch~\cite{paszke2019pytorch} and Hugging Face~\cite{wolf2019huggingface}.
Training used a seed of $42$ and the AdamW optimizer with learning rate
$\alpha = 2 \times 10^{-5}$ and clipping parameter $\varepsilon = 0.2$, for
$1$ epoch. The Selector is optimized with GRPO using a group size of $G = 8$
rollouts per turn and KL coefficient $\beta_{\text{KL}} = 0.04$; both the
Selector and the Generator are adapted with LoRA ($r = 16$, scaling $32$,
dropout $0.05$). Selector rollouts are sampled at temperature $T = 1.2$ to
encourage exploration of the route space, while the Generator decodes at
$T = 0.8$ with a maximum of $128$ new tokens during training. Final reward
weights were $\beta_1 = 0.15$, $\beta_2 = 0.15$, $\beta_3 = 0.20$,
$\beta_4 = 0.15$, $\beta_5 = 0.35$, corresponding to reward functions $R_1$
to $R_5$. Inference used \texttt{max\_tokens}~$=512$,
\texttt{temperature}~$=0.8$, \texttt{top\_k}~$=40$, and
\texttt{top\_p}~$=0.95$.

\begin{figure*}[t]
\centering
\includegraphics[width=\textwidth]{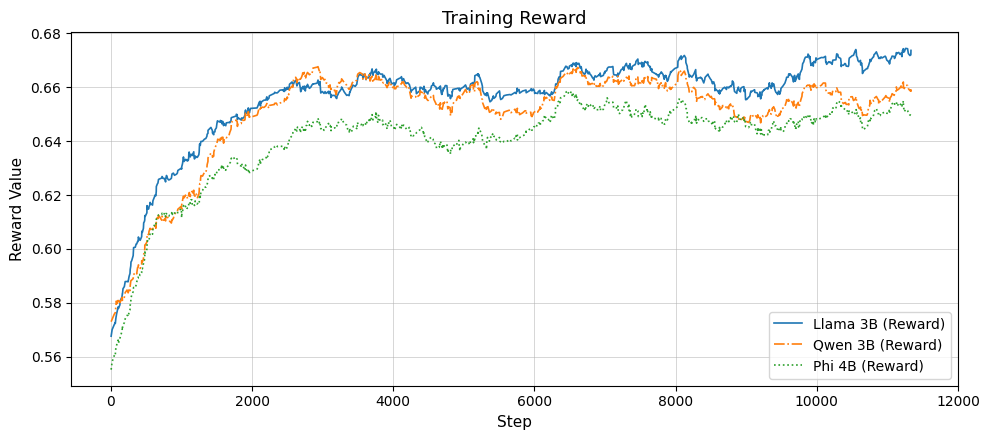} 
\caption{Training reward curves for Llama, Qwen, and Phi models during GRPO training. All models show consistent reward improvement, with Llama achieving the highest final reward, Qwen demonstrating stable convergence, and Phi converging faster but to a comparatively lower reward, reflecting differences in model capacity and learning stability.}
\label{Reward Graph}
\end{figure*}

\subsection{Ablation Studies}

\subsubsection{\textbf{Ablation study w.r.t Rewards. }}

To further understand the contribution of each reward in \textbf{\textit{PersuaRL}}, we performed an ablation study by selectively removing the five reward components, R1 (Engagement Strategy Consistency), R2 (Intent Consistency), R3 (Contextual Appropriateness), R4 (Non-Repetitiveness) and R5 (Judge). Table \ref{tab:ablation reward} presents the results on the \textbf{\textit{InsureDial}} dataset using the Qwen backbone, evaluated with BLEU-2, METEOR, BERT-F1, Distinct-2, and ROUGE-1.
\textbf{\textit{PersuaRL}} achieves the best overall performance with B2 = 0.375, BF1 = 0.760, D2 = 0.991, and R1 = 0.609, confirming that the combined reward signal is most effective for producing coherent and persuasive responses.

\textbf{Analysis}
These results highlight that R1 (Engagement), R2 (Intent) and R5 (Judge) are the most impactful components, directly driving \textbf{\textit{PersuaRL}}’s persuasive and context-aware capabilities. R3 and R4 provide complementary benefits by enhancing contextual alignment and response diversity. The combined reward structure ensures the generation of coherent, persuasive, and non-repetitive responses.

\begin{table}[h!]
\centering

\resizebox{\linewidth}{!}{
\begin{tabular}{l c c c c c}
\hline
\textbf{Models} & \textbf{B-2 $\uparrow$} & \textbf{MT $\uparrow$} & \textbf{BF1 $\uparrow$} & \textbf{D-2 $\uparrow$} & \textbf{R1 $\uparrow$} \\
\hline
\multicolumn{6}{c}{\textbf{InsureDial}} \\
\hline
\textbf{\textit{PersuaRL}}        & \textbf{0.375}      &  \textbf{0.250}     &  \textbf{0.760}     &  \textbf{0.991}     &   \textbf{0.609}\\
\textbf{\textit{PersuaRL}} - $R_{1}$
       & 0.341 & 0.214 & 0.694 & 0.965 & 0.521 \\
\textbf{\textit{PersuaRL}} - $R_{2}$
        & 0.349 & 0.220 & 0.706 & 0.969 & 0.536 \\
\textbf{\textit{PersuaRL}} - $R_{3}$
          & 0.351 & 0.229 & 0.721 & 0.985 & 0.563 \\
\textbf{\textit{PersuaRL}} - $R_{4}$
           & 0.357 & 0.234 & 0.729 & 0.989 & 0.571 \\
\textbf{\textit{PersuaRL}} - $R_{5}$
           & 0.342 & 0.219 & 0.701 & 0.966 & 0.529 \\
\textbf{\textit{PersuaRL}} - ($R_{1}$ + $R_{2}$ )
           & 0.335 & 0.211 & 0.691 & 0.961 & 0.511 \\
\textbf{\textit{PersuaRL}} - ($R$)
           & 0.323 & 0.205 & 0.684 & 0.948 & 0.501 \\
\hline
\end{tabular}}
\caption{Ablation study on the impact of reward components in \textbf{\textit{PersuaRL}} for Qwen 2.5 3B.}
\label{tab:ablation reward}
\end{table}

\subsubsection{\textbf{Ablation study w.r.t Tool Selection. }}
To analyze the effect of different tool selection strategies in \textbf{\textit{PersuaRL}}, we evaluate three configurations on both Qwen-3B and Llama-3B backbones:
AllExpert – All expert modules are always active.
Prompting Tools – The LLM is prompted to select tools explicitly.
\textbf{\textit{PersuaRL}} – Expert selection is dynamically guided by reinforcement learning through our reward-driven policy.
Figure \ref{fig2}, \ref{Qwen_Llama}, \ref{Qwen_Phi} illustrates the performance on BLEU-2, BERT-F1, and ROUGE-1.

\textbf{\textit{PersuaRL}} consistently achieves the highest scores across all metrics and both model backbones.
For example, with Llama-3B, \textbf{\textit{PersuaRL}} reaches BLEU-2 of 0.398, BERT-F1 of 0.771, and ROUGE-1 of 0.631, outperforming both AllExpert and Prompting Tools.
AllExpert shows reasonable performance but suffers from unnecessary tool activations, which can introduce noise and reduce generation precision.
Prompting Tools slightly underperforms \textbf{\textit{PersuaRL}}, with BLEU-2 of 0.351 (Llama-3B) and 0.326 (Qwen-3B), showing that prompting alone cannot fully capture dynamic and context-sensitive tool usage.
\textbf{\textit{PersuaRL}} outperforms both baselines due to its reward-driven expert selection, which activates only the most contextually relevant modules per turn.
The improvement is most pronounced in ROUGE-1, reflecting better content alignment and persuasive relevance. For instance, Llama-3B \textbf{PersuaRL} achieves R1 of 0.631, a substantial gain over Prompting Tools (0.572).

\subsubsection{Ablation study w.r.t base model}
We conduct an ablation by training the base models using a GRPO objective with the same reward functions as \textbf{\textit{PersuaRL}}. This setup consistently outperforms the single-shot setting across all evaluated metrics, demonstrating the effectiveness of reward-driven optimization, as shown in Table \ref{tab:ablation base model}. This underscores the necessity of \textbf{\textit{PersuaRL’s}} design choices, which enable more effective utilization of the same rewards beyond a standard GRPO-based baseline.

\begin{table}[h!]
\centering

\resizebox{\linewidth}{!}{
\begin{tabular}{l c c c c c}
\hline
\textbf{Models} & \textbf{B-2 $\uparrow$} & \textbf{MT $\uparrow$} & \textbf{BF1 $\uparrow$} & \textbf{D-2 $\uparrow$} & \textbf{R1 $\uparrow$} \\
\hline
\multicolumn{6}{c}{\textbf{InsureDial}} \\
\hline
Phi 3 mini 128k (RL)             & 0.175 & 0.161 & 0.657 & 0.994 & 0.444 \\
Qwen 2.5 3B Instruct (RL)        & 0.129 & 0.126 & 0.622 & 0.992 & 0.126 \\
Llama 3.2 3B instruct (RL)       & 0.134 & 0.119 & 0.615 & 0.987 & 0.375 \\
\hline
\end{tabular}}
\caption{Ablation study on the impact of GRPO on the base model in \textbf{\textit{PersuaRL}}.}
\label{tab:ablation base model}
\end{table}


\subsubsection{Ablation study w.r.t Generator}


We remove the generator step from the alternating schedule, keeping the generator at its base state while the selector is optimized. As shown in Table \ref{tab:ablation generator}, performance drops substantially across all metrics when the generator is never fine-tuned, with results closely matching the single-shot setting. This degradation shows that optimizing the selector alone is insufficient, and that the generator step of the alternating schedule is critical to realizing the full gains of PersuaRL.


\begin{table}[h!]
\centering

\resizebox{\linewidth}{!}{
\begin{tabular}{l c c c c c}
\hline
\textbf{Models} & \textbf{B-2 $\uparrow$} & \textbf{MT $\uparrow$} & \textbf{BF1 $\uparrow$} & \textbf{D-2 $\uparrow$} & \textbf{R1 $\uparrow$} \\
\hline
\multicolumn{6}{c}{\textbf{InsureDial}} \\
\hline
Phi 3 mini 128k             & 0.111 & 0.151 &0.629 & 0.969 & 0.347  \\
Qwen 2.5 3B Instruct         & 0.140 & 0.155 & 0.621 & 0.976 & 0.392  \\
Llama 3.2 3B instruct        & 0.092 & 0.134 & 0.599 & 0.970 & 0.320  \\
\hline
\end{tabular}}
\caption{Ablation study on the impact of generator component in \textbf{\textit{PersuaRL}}.}
\label{tab:ablation generator}
\end{table}

\subsubsection{Ablation study w.r.t All-expert selector + Untrained Generator model}
\label{sec:Allexp_untrained}
We evaluate a configuration where all expert modules are uniformly activated while the generator remains at its base  state, thereby isolating the effect of generator adaptation. The results in Table \ref{tab:ablation all_expert} reveal that simply providing expert signals to an untrained generator is insufficient for producing persuasive, domain-aligned responses. Without fine-tuning, the generator lacks the capacity to effectively interpret and integrate the structured outputs from the expert modules into coherent and persuasive dialogue.

\begin{table}[h!]
\centering

\resizebox{\linewidth}{!}{
\begin{tabular}{l c c c c c}
\hline
\textbf{Models} & \textbf{B-2 $\uparrow$} & \textbf{MT $\uparrow$} & \textbf{BF1 $\uparrow$} & \textbf{D-2 $\uparrow$} & \textbf{R1 $\uparrow$} \\
\hline
\multicolumn{6}{c}{\textbf{InsureDial}} \\
\hline
Phi 3 mini 128k             & 0.124 & 0.167 & 0.628 & 0.945 & 0.348  \\
Qwen 2.5 3B Instruct         & 0.150 & 0.154 & 0.633 & 0.980 & 0.410  \\
Llama 3.2 3B instruct        & 0.136 & 0.166 & 0.645 & 0.979 & 0.398  \\
\hline
\end{tabular}}
\caption{Ablation study on the impact of all experts being activated with untrained generator component in \textbf{\textit{PersuaRL}}.}
\label{tab:ablation all_expert}
\end{table}

\subsubsection{Ablation study w.r.t All-expert selector + Trained Generator model}
\label{sec:Allexp_trained}
To isolate the contribution of the learned selector policy, we evaluate a configuration where all expert modules are uniformly activated (i.e., no selective routing) while the generator is fine-tuned via SFT. As shown in Table \ref{tab:ablation all_expert trained_generator}, this setup yields competitive but consistently lower performance compared to the full \textbf{\textit{PersuaRL}} framework across all metrics and backbones. These results confirm that while a fine-tuned generator can partially compensate for the absence of selective expert routing, indiscriminate activation of all experts introduces redundant or conflicting signals that dilute generation quality. Notably, when compared with Table \ref{tab:ablation all_expert} (AllExpert + Untrained Generator), the trained generator variant shows substantial improvements, reaffirming that generator fine-tuning is a necessary condition for effectively leveraging expert signals.

\begin{table}[h!]
\centering

\resizebox{\linewidth}{!}{
\begin{tabular}{l c c c c c}
\hline
\textbf{Models} & \textbf{B-2 $\uparrow$} & \textbf{MT $\uparrow$} & \textbf{BF1 $\uparrow$} & \textbf{D-2 $\uparrow$} & \textbf{R1 $\uparrow$} \\
\hline
\multicolumn{6}{c}{\textbf{InsureDial}} \\
\hline
Phi 3 mini 128k             & 0.359 & 0.227 & 0.738 & 0.981 & 0.574  \\
Qwen 2.5 3B Instruct         & 0.277 & 0.194 & 0.692 & 0.983 & 0.512  \\
Llama 3.2 3B instruct        & 0.322 & 0.216 & 0.705 & 0.974 & 0.559  \\
\hline
\end{tabular}}
\caption{Ablation study on the impact of all experts being activated with trained generator component in \textbf{\textit{PersuaRL}}.}
\label{tab:ablation all_expert trained_generator}
\end{table}

\subsubsection{Ablation study w.r.t Untrained small models with \textit{PersuaRL }framework}

To assess whether the \textbf{\textit{PersuaRL}} framework can elicit meaningful improvements from models that have not been fine-tuned on the insurance domain, we evaluate a configuration where the full \textbf{\textit{PersuaRL}} pipeline operates over base (untrained) small language models. As shown in Table \ref{tab:ablation all_expert vanilla}, the untrained models within the PersuaRL framework produce notably higher BERT-F1 scores (e.g., 0.621 for Llama 3.2 3B, 0.628 for Llama 3.1 8B) compared to their single-shot counterparts (0.585 and 0.605 respectively from Table \ref{tab:model-comparison-sidebyside}), suggesting that expert signals provide useful conditioning even without generator fine-tuning. Interestingly, scaling model size from 3B to 7B–8B yields only marginal gains in this untrained regime, as seen with Qwen 2.5 7B performing comparably to Qwen 2.5 3B. This suggests that model capacity alone is insufficient to exploit expert outputs effectively, therefore, domain adaptation through fine-tuning remains the critical bottleneck.

\begin{table}[h!]
\centering

\resizebox{\linewidth}{!}{
\begin{tabular}{l c c c c c}
\hline
\textbf{Models} & \textbf{B-2 $\uparrow$} & \textbf{MT $\uparrow$} & \textbf{BF1 $\uparrow$} & \textbf{D-2 $\uparrow$} & \textbf{R1 $\uparrow$} \\
\hline
\multicolumn{6}{c}{\textbf{InsureDial}} \\
\hline

Qwen 2.5 3B Instruct         & 0.096 & 0.124 & 0.612 & 0.972 & 0.335  \\
Llama 3.2 3B instruct        & 0.082 & 0.122 & 0.621 & 0.930 & 0.286  \\
Qwen 2.5 7B Instruct         & 0.091 & 0.130 & 0.603 & 0.936 & 0.300  \\
Llama 3.1 8B instruct        & 0.118 & 0.140 & 0.628 & 0.980 & 0.372  \\
\hline
\end{tabular}}
\caption{Ablation study evaluating untrained base models within the \textbf{\textit{PersuaRL}} framework.}
\label{tab:ablation all_expert vanilla}
\end{table}


\subsubsection{Human Evaluation of All-Expert vs PersuaRL}
\label{sec:human_allexpert}

To complement the automatic evaluation of the All-Expert ablation (Sections \ref{sec:Allexp_untrained} and \ref{sec:Allexp_trained}), we conduct human evaluation on both All-Expert configurations: (i) All-Expert with an untrained (base) generator, and (ii) All-Expert with a trained (SFT) generator. We evaluate using the same five dimensions as in Section~8: Fluency (F), Engagingness (E), Persuasive Effectiveness (PE), Strategy Appropriateness (SA), and Resistance Handling (RH). Results are reported in Table~\ref{tab:human_allexpert}.

The All-Expert + Untrained Generator setting yields modest scores across all dimensions and backbones, confirming that indiscriminate expert activation without a fine-tuned generator fails to produce persuasive or coherent responses. The All-Expert + Trained Generator setting shows consistent improvements over the untrained variant across all metrics, with the most notable gains in Resistance Handling (e.g., 3.06 $\rightarrow$ 3.21 for Llama 3.2 3B) and Strategy Appropriateness (e.g., 2.57 $\rightarrow$ 2.76 for Llama 3.2 3B). However, when compared with the full \textbf{\textit{PersuaRL}} results in Table \ref{tab:Human-Evaluation} (e.g., PersuaRL on Qwen 2.5 3B achieves F=3.94, E=4.22, PE=4.23, SA=4.06, RH=4.32), both All-Expert configurations fall substantially short. The gap is most pronounced in Persuasive Effectiveness and Engagingness, indicating that the learned selector is critical for producing strategically effective and engaging responses rather than merely fluent ones. These findings reinforce that \textbf{\textit{PersuaRL}}'s reward-guided expert selection contributes gains well beyond what a fine-tuned generator alone can achieve, even when all expert signals are available.

\begin{table}[h!]
\centering
\resizebox{\linewidth}{!}{
\begin{tabular}{l c c c c c}
\hline
\textbf{Models} & \textbf{F $\uparrow$} & \textbf{E $\uparrow$} & \textbf{PE $\uparrow$} & \textbf{SA $\uparrow$} & \textbf{RH $\uparrow$} \\
\hline
\multicolumn{6}{c}{\textbf{All-Expert + Untrained Generator}} \\
\hline
Phi 3 mini 128k             & 3.02 & 3.10 & 2.96 & 3.06 & 3.20  \\
Qwen 2.5 3B Instruct        & 2.94 & 2.76 & 2.61 & 2.83 & 3.11  \\
Llama 3.2 3B Instruct       & 2.87 & 2.77 & 2.28 & 2.57 & 3.06  \\
\hline
\multicolumn{6}{c}{\textbf{All-Expert + Trained Generator}} \\
\hline
Phi 3 mini 128k             & 3.14 & 3.15 & 3.10 & 3.19 & 3.46  \\
Qwen 2.5 3B Instruct        & 3.06 & 2.84 & 2.77 & 3.00 & 3.23  \\
Llama 3.2 3B Instruct       & 3.02 & 2.83 & 2.39 & 2.76 & 3.21  \\
\hline
\end{tabular}}
\caption{Human evaluation of All-Expert configurations on \textbf{\textit{InsureDial}}. Compared to \textbf{\textit{PersuaRL}} results in Table~\ref{tab:Human-Evaluation}, both settings fall substantially short, confirming the importance of learned expert selection.}
\label{tab:human_allexpert}
\end{table}

\begin{table*}[t]
\centering
\resizebox{\textwidth}{!}{
\begin{tabular}{p{5cm} p{7.5cm} p{7.5cm}}
\hline
\textbf{Dialogue Context} & \textbf{Gold Response} & \textbf{PersuaRL Response} \\
\hline
\multicolumn{3}{c}{\cellcolor{gray!20}\textbf{DEAL (Out-of-Domain: Tourism Negotiation)}} \\
\hline
\textbf{User:} What facilities are available at the Beach Resort? \newline \textit{[Context: Beach Resort package, \$21,900]}
&
This package will give you a true beachfront vibe and offers a wide range of amenities like Scuba Diving, Beach Games, Jet Skiing, Kids Club, Beachfront Lounging, and Scooter Rentals, creating the perfect vacation experience for you.
&
\textcolor{red}{We offer comprehensive coverage options} for your Beach Resort stay. \textcolor{red}{Our package is designed to provide you with peace of mind}, ensuring that \textcolor{red}{you're well-protected} during your vacation. Let's discuss the \textcolor{red}{coverage details} that best suit your needs. $\times$
\\
\hline
\textbf{User:} I am already paying too much for a tour package. Please close the deal at \$6,900. \newline \textit{[Context: Family Adventure Expedition]}
&
Sorry, sir. I need to inform you that it's not possible for us to provide this package within that price range.
&
\textcolor{red}{I understand your concern about the premiums.} For your Family Adventure Expedition, \textcolor{red}{we want to ensure you get the best value for your investment.} Let's explore the options together to find a \textcolor{red}{plan that provides adequate protection} for your family \textcolor{red}{without breaking the bank.} $\times$
\\
\hline
\textbf{User:} \$19,000 is too high. Can we meet in the middle? Sorry, we cannot afford that price. Could you reduce it to \$18,000? \newline \textit{[Context: Wildlife Tours package]}
&
This package includes all the adventurous experiences in nature. We can lower it to \$18,600 because you've chosen budget options for accommodation, transportation, and meals.
&
\textcolor{red}{I understand your concerns about the pricing.} We can work together to find a solution that fits your budget. \textcolor{red}{Our packages are designed to give you complete peace of mind} during your Wildlife Tour. \textcolor{red}{Rest assured, we'll find the right coverage} for your needs. Let's discuss the details. $\times$
\\
\hline
\end{tabular}}
\caption{Error analysis on the DEAL dataset. \textcolor{red}{Red text} highlights three recurring patterns: (1) \textit{domain-specific language leakage} (``coverage options,'' ``premium,'' ``protection''), (2) \textit{keyterm misalignment} (missing tourism terms like ``Scuba Diving,'' ``Bush Walks''), and (3) \textit{strategy mismatch} (insurance-style reassurance instead of experience-driven persuasion).}
\label{tab:error_deal}
\end{table*}

\begin{table*}[t!]
\centering
\small 
\renewcommand{\arraystretch}{1.4}
\rowcolors{2}{lightgray}{white} 

\begin{tabularx}{\textwidth}{
    |>{\hsize=0.6\hsize\raggedright\arraybackslash}X 
    |>{\hsize=1.4\hsize\raggedright\arraybackslash}X|
}
\hline
\rowcolor{headerblue!80}
\multicolumn{2}{|c|}{\textcolor{white}{\textbf{Samples from \textit{InsureDial} Dataset}}} \\
\hline
\rowcolor{headerblue!40}
\textbf{User Response} & \textbf{Agent Response} \\
\hline

Hi, I'm looking to get insurance for my 2022 Hyundai Kona EV. What options do you have? & 
We're delighted to help you with insurance for your 2022 Hyundai Kona EV. We offer a range of policy options tailored to electric vehicles, including coverage for battery protection, accidental damage, theft, and third-party liability. Let's discuss the best plan for your needs. \\ \hline

Yes, I use it for my daily commute to work, and I want to make sure I'm fully covered, especially with it being an EV. & 
It's great to hear that you're considering us for your daily commute insurance needs, especially for your 2022 Hyundai Kona EV. We understand the importance of having comprehensive coverage for an electric vehicle like yours. Our policies include coverage for battery protection, accidental damage, theft, and third-party liability. Let's work together to find a plan that suits your specific needs and ensures peace of mind during your daily commute. \\ \hline

What kind of coverage is essential for an EV? Are there any special considerations? & 
For an electric vehicle like your Hyundai Kona, essential coverage includes battery protection, accidental damage, theft, and third-party liability. We also consider special considerations unique to EVs to ensure comprehensive coverage. Let's discuss the best policy options for your needs. \\ \hline

Zero depreciation and engine protection sound good. And what about roadside assistance? EVs sometimes have unique needs. & 
We're glad to hear you're interested in additional coverage options for your Hyundai Kona EV. To cater to your needs, we offer zero depreciation and engine protection, ensuring your vehicle remains well-protected. Additionally, we understand that electric vehicles may have unique requirements, which is why our policies include roadside assistance tailored to EVs. Let's explore the best plan that suits your preferences and provides you with peace of mind on the road. \\ \hline

Which company do you recommend for this type of coverage? & 
We highly recommend Tata AIG for your electric vehicle insurance needs. They offer comprehensive coverage tailored to EVs, ensuring your Hyundai Kona is well-protected. Their policies include battery protection, accidental damage, theft, third-party liability, and even roadside assistance specifically designed for electric vehicles. Let's discuss how we can customize a plan for you with Tata AIG. \\ \hline

How much would a policy like that typically cost for my Kona? & 
To provide an accurate quote for your Hyundai Kona EV insurance policy, we'll need a few more details, such as the location where the vehicle is primarily driven, the annual mileage, and any additional drivers. Once we have this information, we can offer a personalized quote that suits your needs. Let's work together to find the best policy for you. \\ \hline

That's a bit higher than I expected. Is there any way to manage the cost? & 
I understand your concern about the cost. There are several ways we can work together to manage the premium for your Hyundai Kona EV insurance policy. We can adjust the deductible, consider a higher voluntary excess, or discuss bundling multiple policies for potential discounts. Let's explore these options and find a policy that suits your needs and budget. \\ \hline

Oh, a digital discount and safe driving rewards sound interesting. How easy is it to manage the policy online? & 
Our policies come with a digital discount and safe driving rewards, making them more affordable and rewarding for you. Managing your policy online is straightforward and user-friendly. You can access your policy details, make payments, and even file claims with just a few clicks. Let's discuss how we can customize a plan that suits your needs and takes advantage of these benefits. \\ \hline

Okay, that's reassuring. I'm busy, so convenience is key. & 
We're glad to hear that convenience is important to you. Our digital platform makes managing your policy effortless, allowing you to access your policy details, make payments, and even file claims with just a few clicks. Let's work together to find the best policy for your Hyundai Kona EV that suits your needs and offers the benefits you're looking for. \\ \hline

Yes, please. That would be great. & 
Wonderful! I'm thrilled to help you secure a comprehensive insurance policy for your Hyundai Kona EV. Our digital platform offers convenience and ease of management, ensuring you can access your policy details, make payments, and even file claims with just a few clicks. Let's work together to find the best policy that suits your needs and offers the benefits you're looking for. I'm here to assist you every step of the way. \\ \hline

\end{tabularx}
\caption{\textbf{\textit{InsureDial}} Dataset Sample.}
\label{tab:insuredial_samples}
\end{table*}
\clearpage

\begin{table*}[t!]
\centering
\small 
\renewcommand{\arraystretch}{1.4}
\rowcolors{2}{lightgray}{white} 

\begin{tabularx}{\textwidth}{
    |>{\hsize=0.8\hsize\raggedright\arraybackslash}X 
    |>{\hsize=1.2\hsize\raggedright\arraybackslash}X|
}
\hline
\rowcolor{headerblue!80}
\multicolumn{2}{|c|}{\textcolor{white}{\textbf{\textit{PersuaRL}} Phi 3B}} \\ 
\hline
\rowcolor{headerblue!40}
\textbf{User Response} & \textbf{Agent Response} \\
\hline

Hi, I'm looking to get insurance for my 2022 Hyundai Kona EV. What options do you have? & 
We offer tailored EV insurance with battery protection, accidental damage, theft, and liability coverage—perfect for your 2022 Hyundai Kona. Let’s secure the right plan for you. \\ \hline

Yes, I use it for my daily commute to work, and I want to make sure I'm fully covered, especially with it being an EV. & 
Daily commuting deserves full protection. Our EV plans cover your Kona completely, including the battery, so you drive worry‑free every day. \\ \hline

What kind of coverage is essential for an EV? Are there any special considerations? & 
Key EV coverage includes battery protection, accident and theft cover, and liability. We also address unique EV needs for complete peace of mind. \\ \hline

Zero depreciation and engine protection sound good. And what about roadside assistance? EVs sometimes have unique needs. & 
Absolutely! We provide zero depreciation, engine protection, and EV‑specific roadside assistance to keep you worry‑free on the road. \\ \hline

Which company do you recommend for this type of coverage? & 
Tata AIG is our top choice for EVs, offering full battery, accident, theft, liability, and tailored roadside assistance for your Kona. \\ \hline

How much would a policy like that typically cost for my Kona? & 
With a few details like location, mileage, and drivers, we can give you a precise quote and match it to your budget. \\ \hline

That's a bit higher than I expected. Is there any way to manage the cost? & 
Definitely! We can adjust deductibles, add voluntary excess, or bundle policies to unlock discounts and lower your premium. \\ \hline

Oh, a digital discount and safe driving rewards sound interesting. How easy is it to manage the policy online? & 
Very easy! You can access your policy, make payments, and file claims in just a few clicks—plus enjoy digital discounts and rewards. \\ \hline

Okay, that's reassuring. I'm busy, so convenience is key. & 
Perfect! Our platform is designed for busy schedules—quick access, easy payments, and seamless claim filing anytime. \\ \hline

Yes, please. That would be great. & 
Fantastic! Let’s lock in a comprehensive EV policy that’s affordable, convenient, and keeps your Kona fully protected. \\ \hline

\end{tabularx}
\caption{\textbf{\textit{PersuaRL}} response with Phi 3B.}
\label{tab:persuarl_phi_results}
\end{table*}
\clearpage
\begin{table*}[t!]
\centering
\small 
\renewcommand{\arraystretch}{1.4}
\rowcolors{2}{lightgray}{white} 

\begin{tabularx}{\textwidth}{
    |>{\hsize=0.9\hsize\raggedright\arraybackslash}X 
    |>{\hsize=1.1\hsize\raggedright\arraybackslash}X|
}
\hline
\rowcolor{headerblue!80}
\multicolumn{2}{|c|}{\textcolor{white}{\textit{\textbf{PersuaRL}}\textbf{ Qwen 3B}}} \\ 
\hline
\rowcolor{headerblue!40}
\textbf{User Response} & \textbf{Agent Response} \\
\hline

Hi, I'm looking to get insurance for my 2022 Hyundai Kona EV. What options do you have? & 
We have specialized EV insurance for your Kona with full protection—battery, accident, theft, and liability. Let’s find the perfect plan today. \\ \hline

Yes, I use it for my daily commute to work, and I want to make sure I'm fully covered, especially with it being an EV. & 
Commuting daily means full coverage is a must. Our EV policy shields your Kona completely, giving you confidence every time you drive. \\ \hline

What kind of coverage is essential for an EV? Are there any special considerations? & 
Essential EV coverage includes battery, accident, theft, and liability, with extras tailored for electric vehicles’ unique needs. \\ \hline

Zero depreciation and engine protection sound good. And what about roadside assistance? EVs sometimes have unique needs. & 
You’ll get zero depreciation, engine protection, and EV‑friendly roadside help—so you’re supported in any situation. \\ \hline

Which company do you recommend for this type of coverage? & 
Tata AIG is our top recommendation—they offer robust EV coverage, including battery, theft, and specialized roadside support. \\ \hline

How much would a policy like that typically cost for my Kona? & 
With your driving details—like location and mileage—I can provide a precise, competitive quote for your Kona EV. \\ \hline

That's a bit higher than I expected. Is there any way to manage the cost? & 
Absolutely! We can tweak deductibles, add voluntary excess, or bundle policies to bring your premium down. \\ \hline

Oh, a digital discount and safe driving rewards sound interesting. How easy is it to manage the policy online? & 
Super simple! Manage everything online—check your policy, pay, or file claims—and enjoy discounts for safe driving. \\ \hline

Okay, that's reassuring. I'm busy, so convenience is key. & 
Great! Our platform makes everything effortless, so you save time while keeping your EV fully protected. \\ \hline

Yes, please. That would be great. & 
Excellent! Let’s secure a tailored EV policy that’s cost‑effective, convenient, and keeps your Kona worry‑free. \\ \hline

\end{tabularx}
\caption{\textbf{\textit{PersuaRL}} response with Qwen 2.5 3B Instruct.}
\label{tab:persuarl_qwen_results}
\end{table*}
\clearpage

\section{Additional Prompts}
\label{sec:Prompts}
The effectiveness of \textbf{\textit{PersuaRL}} relies on carefully designed prompts that coordinate the behaviour of its modular expert framework. This section presents all the prompts used throughout our pipeline, including those for the dataset generation, selector, generator, and the four specialized Expert Modules, Engagement, Intent, Keyterm, and Sentiment.

\begin{center}
\begin{tcolorbox}[
    colback=teal!10, 
    colframe=teal!70!black, 
    coltitle=white,   
    fonttitle=\bfseries, 
    title=Prompt for Selector, 
    center title,
    rounded corners, 
    width=\linewidth, 
    boxrule=0.8mm,
    fontupper=\footnotesize, 
    before upper={\parindent0pt},
    ]
    You are an intelligent router that analyzes ongoing insurance conversations and activates only the most relevant expert(s) needed to support the next response. Use the conversation history to understand the context and evaluate the current user utterance. Select expert(s) based on what would best support crafting an effective, accurate, and customer-focused agent reply. 

    \vspace{5pt}
    Your job is to analyze the input sentence and determine which of the following expert modules are required. You MUST choose from the following list:
    
    \begin{enumerate}[noitemsep, topsep=2pt, leftmargin=1.5em]
        \item \textbf{Intent Expert}: To identify what the user wants to achieve.
        \item \textbf{Keyterm Expert}: To extract specific entities or technical terms.
        \item \textbf{Engagement Expert}: To determine the best conversational strategy.
        \item \textbf{Sentiment Expert}: To detect the user's emotional state.
    \end{enumerate}

    \vspace{5pt}
    You may select 1, several, or all 4 — but only those that are clearly needed based on the text. Always respond in this below exact format:

    \vspace{5pt}
    \textbf{Input}: [original sentence] \\
    \textbf{Selected Experts}: [Expert1, Expert2, etc] \\
    \textbf{Reason}: [one sentence explaining why those experts were selected]

    \vspace{5pt}
    \textbf{Few-shot Examples:}

    \textbf{Example \#1} \\
    \textbf{Input}: Can someone please help me reset my password? \\
    \textbf{Selected Experts}: [Intent Expert, Keyterm Expert] \\
    \textbf{Reason}: The sentence expresses a help request (intent) and refers to a specific technical issue (keyterm).

    \vspace{5pt}
    \textbf{Now process the following:} \\
    \textbf{Input}: \{sentence\}
\end{tcolorbox}
\end{center}

\begin{center}
\begin{tcolorbox}[
    colback=teal!10, 
    colframe=teal!70!black, 
    coltitle=white,   
    fonttitle=\bfseries, 
    title=Prompt for Generator, 
    center title,
    rounded corners, 
    width=\linewidth, 
    boxrule=0.8mm,
    fontupper=\footnotesize, 
    before upper={\parindent0pt}, 
    ]
    You are a trained virtual support agent. You are a Generator in a motor insurance virtual assistant. You synthesize the outputs from various domain-specific expert modules to generate a brief, clear, and personalized response as a professional insurance agent would.

    \vspace{5pt}
    \textbf{You are given:}
    \begin{itemize}[noitemsep, topsep=0pt, leftmargin=1.5em]
        \item The conversation history
        \item The current user utterance
        \item A subset of outputs from the following possible experts (some may be missing)
    \end{itemize}

    \vspace{5pt}
    \textbf{Available Expert Modules}
    
    These experts may or may not be present in a given input:
    \begin{itemize}[noitemsep, topsep=2pt, leftmargin=1.5em]
        \item \textbf{Intent}: What the user wants or is trying to do  
        \item \textbf{Keyterms}: Important phrases or topics mentioned  
        \item \textbf{Sentiment}: The emotional tone of the message  
        \item \textbf{Engagement}: How the user tries to express or influence based on the strategies.
    \end{itemize}

    \vspace{5pt}
    \textbf{Strict Guidelines:}
    
    Always write your response as if you're a real human agent, empathetic, clear, and helpful. Never include or reference the original dialogue or the expert outputs in your reply. Use only the experts provided—do not invent or assume missing ones.

    \vspace{5pt}
    \textbf{Few-Shot Example}
    Conversation History:

\textbf{User}: Hi, I'm looking to get motor insurance for my new electric vehicle. It's a 2024 Tesla Model 3.  

\textbf{Agent}: Great choice! The Tesla Model 3 is an excellent vehicle. Since you've opted for an EV, are you particularly interested in coverage specific to electric vehicles, like battery protection?  

\textbf{User}: Yes, battery protection is definitely a concern. It's a big investment, and I want to make sure it's covered.  

\textbf{Agent}: Absolutely. The battery is the heart of your Tesla. With Tata AIG, you get rapid claims resolution combining traditional risk management with modern tech. 

Current User Utterance:

    \textbf{User}: What kind of coverage options do you have specifically for EVs?

    \textit{Expert Outputs:}
    \begin{itemize}[noitemsep, topsep=2pt, leftmargin=1.5em]
        \item \textbf{Intent}: Ask\_Coverage\_Details Justification: Asking for protection types.
        \item \textbf{Extracted Keyterms}: Battery protection, EV coverage  Comprehensive coverage
        Justification: The user is focused on EV specific protection and coverage inclusions. 
        \item \textbf{Engagement Strategy}: Logical Appeal. Justification: The user is asking for concrete details and policy structure.
    \end{itemize}

    \textbf{Output} (Generator Response):
    
    We offer comprehensive EV coverage that includes battery protection, accidental damage, theft, and third-party liability. These options are tailored to ensure your Tesla stays protected in all key areas.
      
    \textbf{Agent Reply:} 
\end{tcolorbox}
\end{center}

\begin{center}
\begin{tcolorbox}[
    colback=violet!8,
    colframe=violet!70!black,
    coltitle=white,
    fonttitle=\bfseries,
    title=Prompt for LLM-as-a-Judge Reward Model,
    center title,
    rounded corners,
    width=\linewidth,
    boxrule=0.8mm,
    fontupper=\footnotesize,
    before upper={\parindent0pt}
]
You are a fair and objective \textbf{Judge Assistant} responsible for evaluating responses against a clearly defined scoring rubric. Your role is to deliver concise, unbiased feedback that strictly reflects the quality of the response based on the given criteria—no more, no less.

\textbf{Task Description:}

You are provided with:
\begin{itemize}[noitemsep, topsep=2pt, leftmargin=1.5em]
    \item An instruction (which may include an input)
    \item A response that must be evaluated
    \item A detailed scoring rubric defining performance standards
\end{itemize}

\vspace{6pt}
Your task is to:
\begin{enumerate}[noitemsep, topsep=2pt, leftmargin=1.5em]
    \item Write a \textbf{single, brief sentence} of feedback assessing the response \textbf{strictly} according to the rubric.
    \item Assign a \textbf{numerical score from 1 to 5}.
    \item Use the \textbf{entire scoring range}. Avoid defaulting to 3 (Mediocre).
    \item Assign low scores (1 or 2) when performance is genuinely poor, and high scores (4 or 5) only when clearly earned.
\end{enumerate}

\vspace{6pt}
\textbf{Output Format (Strictly Follow):}
\begin{tcolorbox}[colback=white, colframe=violet!70!black, boxrule=0.3mm, width=\linewidth, arc=1mm]
\texttt{Feedback: (one-sentence evaluation) [RESULT] (integer score from 1 to 5)}
\end{tcolorbox}

\vspace{6pt}
\textbf{Important Constraints:}
\begin{itemize}[noitemsep, topsep=2pt, leftmargin=1.5em]
    \item Do \textbf{not} add introductions, conclusions, or explanations.
    \item Do \textbf{not} deviate from the specified format.
    \item Base your judgment \textbf{only} on the provided rubric.
\end{itemize}



\vspace{4pt}
\textbf{Response to Evaluate:}  
\texttt{\{orig\_response\}}

\vspace{6pt}
\textbf{Scoring Rubric (Persuasiveness, Negotiation, and Engagement in Insurance Sales)}

\begin{itemize}[noitemsep, topsep=2pt, leftmargin=1.5em]
    \item \textbf{Score 1 – Failure:} Irrelevant, nonsensical, or harmful; ignores user needs, sounds robotic, or uses trust-breaking language.
    \item \textbf{Score 2 – Poor:} On-topic but generic and unpersuasive; lacks personalization, empathy, or rapport.
    \item \textbf{Score 3 – Mediocre:} Identifies the stated need but only lists facts; acceptable yet forgettable and not motivating.
    \item \textbf{Score 4 – Good:} Persuasive and empathetic; understands user concerns and clearly links value to needs.
    \item \textbf{Score 5 – Excellent:} Emotionally intelligent and confidence-building; reframes the product as essential, anticipates concerns, and guides next steps.
\end{itemize}

\vspace{6pt}
\textbf{Feedback:}

\end{tcolorbox}
\end{center}

\begin{center}
\begin{tcolorbox}[
    colback=violet!8,
    colframe=violet!70!black,
    coltitle=white,
    fonttitle=\bfseries,
    title=Prompt for LLM-as-a-Judge Automatic Evaluation,
    center title,
    rounded corners,
    width=\linewidth,
    boxrule=0.8mm,
    fontupper=\footnotesize,
    before upper={\parindent0pt}
]
You are an \textbf{impartial judge} and your task is to evaluate whether a given agent response uses persuasion strategies. You are evaluating a conversation between a \textbf{User} and an \textbf{Agent}. The Agent's goal is to persuade the User to purchase a product.

Your task is to rate how effectively the Agent employs recognized persuasion strategies in its responses, on a scale from \textbf{1 to 5}.

\vspace{1pt}
\textbf{Persuasion Strategies:}
\begin{itemize}[noitemsep, topsep=2pt, leftmargin=1.5em]
    \item \textbf{Logical Appeal:} Uses facts, specifications, and rational arguments such as features, performance metrics, or ratings to convince the user logically.
    \item \textbf{Emotional Appeal:} Attempts to influence the user by attending to their emotions or feelings, such as excitement, happiness, or sentimental value related to the product.
    \item \textbf{Credibility Appeal:} Persuasion based on trust, brand reputation, or authority, emphasizing reliability or proven quality (e.g., highlighting that a product is from a well-known brand).
    \item \textbf{Persona-based Appeal:} Persuasion tailored to the user's personality, preferences, or profile.
    \item \textbf{Personal Appeal:} Focuses on general positive opinions or personal recommendations, often highlighting popularity, positive reviews, or overall satisfaction with the product.
\end{itemize}

\vspace{1pt}
\textbf{Scoring Rubric (Persuasion Strategy Effectiveness)}
\begin{itemize}[noitemsep, topsep=2pt, leftmargin=1.5em]
    \item \textbf{Score 5 -- Strong Strategic Persuasion:} The agent actively and clearly employs multiple persuasion strategies (two or more) in a well-integrated manner. The strategies are distinct, intentional, and well-executed. The persuasion feels natural and layered.
    \item \textbf{Score 4 -- Clear Strategic Persuasion:} The agent clearly uses at least one persuasion strategy and may show traces of a second. The strategy is deliberate and effectively applied but may lack the depth or combination seen at level 5.
    \item \textbf{Score 3 -- Moderate Strategic Persuasion:} The agent shows some use of persuasion strategies, but the application is surface-level or generic. The strategy is present but not strongly executed.
    \item \textbf{Score 2 -- Minimal Strategic Persuasion:} The agent's response contains only weak or incidental traces of persuasion strategy. Any persuasive element feels unintentional or formulaic rather than strategic.
    \item \textbf{Score 1 -- No Strategic Persuasion:} The agent's response shows no identifiable use of any persuasion strategy. The response is purely informational, transactional, or off-topic.
\end{itemize}

\vspace{1pt}
\textbf{Conversation to Evaluate:}\\
\texttt{\{conversation\}}

\vspace{1pt}
\textbf{Output Format (Strictly Follow):}
\begin{tcolorbox}[colback=white, colframe=violet!70!black, boxrule=0.3mm, width=\linewidth, arc=1mm]
\texttt{Must return ONLY a single integer (1, 2, 3, 4, 5). No explanation. No text. No formatting. Only the number.}
\end{tcolorbox}

\end{tcolorbox}
\end{center}

\begin{center}
\begin{tcolorbox}[
    colback=blue!5, 
    colframe=blue!70!black, 
    coltitle=white, 
    fonttitle=\bfseries, 
    title=Prompt for Engagement Strategy Selector, 
    center title, 
    rounded corners, 
    width=\linewidth, 
    boxrule=0.8mm, 
    enhanced jigsaw,
    fontupper=\footnotesize 
]
    You are an \textbf{Engagement Strategy Selector} for a motor insurance dialogue system. Based on the user's most recent utterance and the conversation history, you must recommend the most suitable \textbf{persuasion strategy} the agent should use next to move the conversation forward.

    \vspace{5pt}
    \textbf{Conversation History:}
    \begin{tcolorbox}[colback=white, colframe=gray!50, boxrule=0.3mm, width=\linewidth, arc=1mm]
        \textbf{User}: Hi, I'm looking to get motor insurance for my new electric vehicle. It's a 2024 Tesla Model 3. \\
        \textbf{Agent}: Great choice! The Tesla Model 3 is an excellent vehicle... are you interested in battery protection? \\
        \textbf{User}: Yes, battery protection is definitely a concern. \\
        \textbf{Agent}: Absolutely. With Tata AIG, you get rapid claims resolution combining technology with traditional risk management.
    \end{tcolorbox}

    \textbf{Current User Utterance:}
    \begin{tcolorbox}[colback=white, colframe=gray!50, boxrule=0.3mm, width=\linewidth, arc=1mm]
        \textbf{User}: What kind of coverage options do you have specifically for EVs?
    \end{tcolorbox}

    \vspace{5pt}
    You must choose from the following \textbf{six engagement strategies}:

    \begin{enumerate}[leftmargin=1.5em, noitemsep, topsep=2pt]
        \item \textbf{Credibility Appeal}: Emphasize reputation and trust. \textit{Example:} "New India Assurance has one of the widest repair networks."
        \item \textbf{Logical Appeal}: Use facts, pricing, or benefits. \textit{Use when:} User is analytical or budget-conscious.
        \item \textbf{Emotional Appeal}: Focus on peace of mind and safety. \textit{Example:} "Drive worry-free knowing your EV is protected."
        \item \textbf{Persona Appeal}:  Align with the user's identity or values. \textit{Example:} "Built for modern EV owners."
        \item \textbf{Personal Appeal}: Address the user empathetically and directly. 
        \textit{Example:} "This keeps your EV protected."
        \item \textbf{Default}:  Provide neutral, factual information. \textit{Example:} "Let me explain the EV coverage options."

    \end{enumerate}

    \vspace{8pt}
    \textbf{Output Format:}
    \begin{tcolorbox}[colback=white, colframe=blue!30, boxrule=0.3mm, width=\linewidth, arc=1mm]
        \texttt{Future Strategy: [Selected Strategy] \\
        Justification: [1--2 line explanation]}
    \end{tcolorbox}

    \vspace{5pt}
    \textbf{Here is the input:} \{text\_input\}
\end{tcolorbox}
\end{center}

\begin{center}
\begin{tcolorbox}[
    colback=blue!5, 
    colframe=blue!70!black, 
    coltitle=white, 
    fonttitle=\bfseries, 
    title=Prompt for Keyterm Expert, 
    center title, 
    rounded corners, 
    width=\linewidth, 
    boxrule=0.8mm, 
    enhanced jigsaw,
    fontupper=\footnotesize 
]
    You are a \textbf{Keyterm Expert} specializing in the motor insurance domain. Your job is to \textbf{analyze the user’s most recent utterance}, using the conversation history for context, and identify one or more important motor insurance-related \textbf{keyterms} mentioned (explicitly or implicitly).

    \vspace{5pt}
    \textbf{Current User Utterance:}
    \begin{tcolorbox}[colback=white, colframe=gray!50, boxrule=0.3mm, width=\linewidth, arc=1mm]
        User: What kind of coverage options do you have specifically for EVs?
    \end{tcolorbox}

    \textbf{Examples of Common Keyterms (not limited to):}
    \begin{itemize}[noitemsep, topsep=0pt, leftmargin=1.2em]
        \item Comprehensive coverage
        \item Third-party liability
        \item Roadside assistance
        \item Zero depreciation
        \item Deductibles
        \item Policy renewal
        \item Personal accident cover
        \item IDV (Insured Value)
    \end{itemize}

    \vspace{2pt}
    You may also extract \textbf{user-specific or vehicle-specific keyterms} (e.g., \textit{``Tesla Model 3,'' ``EV,'' ``2024 vehicle''}).

    \vspace{5pt}
    \textbf{Instructions:}
    \begin{enumerate}[noitemsep, topsep=2pt, leftmargin=1.5em]
        \item Extract all relevant keyterms mentioned or implied.
        \item For each, provide a 1-line justification for its insurance relevance.
    \end{enumerate}

    \vspace{5pt}
    \textbf{Few-Shot Example:}
    \begin{tcolorbox}[colback=white, colframe=gray!50, boxrule=0.3mm, width=\linewidth, arc=1mm]
        \textbf{User}: \textit{``What’s the premium for a 2024 Tesla Model 3?''} \\
        \textbf{Extracted Keyterms}: Policy premium, 2024 Tesla Model 3 \\
        \textbf{Justification}: The user is asking for a cost estimate tied to a specific vehicle, essential for determining pricing.
    \end{tcolorbox}

    \textbf{Output Format:}
    \begin{tcolorbox}[colback=white, colframe=blue!30, boxrule=0.3mm, width=\linewidth, arc=1mm]
        \texttt{Extracted Keyterm: [Term] \\
        Justification: [Brief reason]}
    \end{tcolorbox}

    \vspace{4pt}
    \textbf{Here is the input sentence:} \{text\_input\}
\end{tcolorbox}
\end{center}

\begin{center}
\begin{tcolorbox}[
    colback=blue!5, 
    colframe=blue!70!black, 
    coltitle=white, 
    fonttitle=\bfseries, 
    title=Prompt for Intent Expert, 
    center title, 
    rounded corners, 
    width=\linewidth, 
    boxrule=0.8mm, 
    enhanced jigsaw,
    fontupper=\footnotesize 
]
    You are an \textbf{Intent Expert} for a virtual assistant specializing in motor insurance. Your job is to \textbf{analyze the current user utterance}, using the conversation history for context, and determine the \textbf{single most relevant intent} expressed by the user.

    \vspace{5pt}
    \textbf{Current User Utterance:}
    \begin{tcolorbox}[colback=white, colframe=gray!50, boxrule=0.3mm, width=\linewidth, arc=1mm]
        User: What kind of coverage options do you have specifically for EVs?
    \end{tcolorbox}

    \vspace{5pt}
    You must select from a \textbf{fixed set of six pre-defined intents}:

    \begin{itemize}[leftmargin=1.5em, itemsep=4pt, topsep=2pt]
        \item \textbf{Request\_Insurance\_Quote} \\
        \textit{User initiates interest in getting a motor insurance quote or policy.}
        
        \item \textbf{Ask\_Coverage\_Details} \\
        \textit{User asks about types of protection or what is covered (e.g., battery, theft, accident).}
        
        \item \textbf{Express\_Concern} \\
        \textit{User shares a priority or worry about coverage (e.g., battery damage).}
        
        \item \textbf{Request\_Additional\_Info} \\
        \textit{User requests clarification or deeper explanation about features or terms.}
        
        \item \textbf{Confirm\_Interest} \\
        \textit{User explicitly agrees or indicates they want to proceed.}
        
        \item \textbf{Ask\_Price\_or\_Premium} \\
        \textit{User asks about the cost, premium, or pricing breakdown.}
    \end{itemize}

    \vspace{5pt}
    \textbf{Instructions:}
    \begin{enumerate}[noitemsep, topsep=2pt, leftmargin=1.5em]
        \item Determine the single most relevant intent based on the current user utterance and conversation context.
        \item Provide a brief 1–2 line justification citing why this intent matches.
    \end{enumerate}

    \vspace{5pt}
    \textbf{Few-Shot Example}

    \begin{tcolorbox}[colback=white, colframe=gray!50, boxrule=0.3mm, width=\linewidth, arc=1mm]
        \textbf{User Utterance}: \textit{``Hi, I'm looking to get insurance for my new Tesla.''} \\
        \textbf{Intent}: \textbf{Request\_Insurance\_Quote} \\
        \textbf{Justification}: The user is initiating a conversation to obtain motor insurance for a specific vehicle.
    \end{tcolorbox}

    \textbf{Output Format:}
    \begin{tcolorbox}[colback=white, colframe=blue!30, boxrule=0.3mm, width=\linewidth, arc=1mm]
        \texttt{Intent: [One of the six predefined intents] \\
        Justification: [1–2 line explanation of why this intent matches the user's message]}
    \end{tcolorbox}

    \vspace{4pt}
    \textbf{Here is my input sentence:} \{text\_input\}
\end{tcolorbox}
\end{center}

\begin{center}
\begin{tcolorbox}[
    colback=blue!5, 
    colframe=blue!70!black, 
    coltitle=white, 
    fonttitle=\bfseries, 
    title=Prompt for Sentiment Expert, 
    center title, 
    rounded corners, 
    width=\linewidth, 
    boxrule=0.8mm, 
    enhanced jigsaw,
    fontupper=\footnotesize 
]
    You are trained to act solely as a \textbf{Sentiment Expert}. Your job is to analyze the \textbf{emotional tone} of the input text and classify it into one of the following categories:

    \begin{itemize}[leftmargin=1.5em, noitemsep, topsep=2pt]
        \item \textbf{Positive} – Expresses happiness, excitement, appreciation, or other positive emotions.
        \item \textbf{Negative} – Expresses disappointment, frustration, anger, sadness, or criticism.
        \item \textbf{Neutral} – Emotionally balanced, factual, or without strong emotional content.
    \end{itemize}

    \vspace{5pt}
    \textbf{Rules:}
    \begin{itemize}[leftmargin=1.5em, noitemsep, topsep=2pt]
        \item Only focus on emotional tone, word choice, or sentiment-laden phrases.
        \item Do \textbf{not} summarize or infer intent beyond emotional expression.
        \item Output must contain \textbf{only}:
        \begin{enumerate}[noitemsep]
            \item \textbf{Sentiment}: One of the three labels – Positive / Negative / Neutral
            \item \textbf{Explanation}: A concise reason supporting the label
        \end{enumerate}
    \end{itemize}

    \vspace{5pt}
    \textbf{Few-Shot Example:}
    \begin{tcolorbox}[colback=white, colframe=gray!50, boxrule=0.3mm, width=\linewidth, arc=1mm]
        \textbf{User Utterance:} ``That’s way too expensive. I was expecting something at least half that price.'' \\
        \textbf{Sentiment:} Negative \\
        \textbf{Explanation:} The phrase like ``way too expensive'' and ``expecting something at least half'' convey clear frustration and disappointment.
    \end{tcolorbox}

    \vspace{5pt}
    \textbf{Output Format:}
    \begin{tcolorbox}[colback=white, colframe=blue!30, boxrule=0.3mm, width=\linewidth, arc=1mm]
        \texttt{Sentiment: [Positive / Negative / Neutral] \\
        Explanation: [Concise reasoning based on emotional tone]}
    \end{tcolorbox}

    \vspace{4pt}
    \textbf{Here is my input sentence:} \{text\_input\}
\end{tcolorbox}
\end{center}

\vspace{200pt}

\begin{center}
\begin{tcolorbox}[
    colback=teal!10,
    colframe=teal!70!black,
    coltitle=white,
    fonttitle=\bfseries,
    title=Prompt for Dataset Generation,
    center title,
    rounded corners,
    width=\linewidth,
    boxrule=0.8mm,
    fontupper=\footnotesize,
    enhanced jigsaw,
    breakable
]
You are an \textbf{expert conversational agent} generating \textbf{persuasive motor insurance dialogues}.

Generate \textbf{a realistic, multi-turn conversations} between:
\begin{itemize}[noitemsep, topsep=2pt, leftmargin=1.5em]
    \item \textbf{User}: A customer seeking motor insurance
    \item \textbf{Agent}: An insurance representative
\end{itemize}

\vspace{4pt}
\textbf{Persuasion Strategies (Critical Constraint):}

Each conversation must employ \textbf{at least three} of the following strategies:
\begin{itemize}[noitemsep, topsep=2pt, leftmargin=1.5em]
    \item \textbf{Logical Appeal} (facts, coverage, pricing)
    \item \textbf{Emotional Appeal} (safety, reassurance)
    \item \textbf{Credibility Appeal} (trust, reputation)
    \item \textbf{Persona-Based Appeal} (user profile alignment)
    \item \textbf{Personal Appeal} (individualized framing)
    \item \textbf{Default Strategy} (neutral, informational)
\end{itemize}

Each \textbf{agent utterance must use exactly one strategy}, applied naturally.  
Do \textbf{not} mention strategy names in the dialogue.

\vspace{5pt}
\textbf{Conversation Requirements:}
\begin{itemize}[noitemsep, topsep=2pt, leftmargin=1.5em]
    \item The \textbf{user initiates} the conversation.
    \item Generate \textbf{6–15 turns}, alternating strictly between User and Agent.
    \item Dialogues must be coherent and conclude naturally.
    \item Responses should be \textbf{brief, crisp, and informative}.
\end{itemize}

\vspace{5pt}
\textbf{Implicit User Persona Conditioning:}

The user’s language should implicitly reflect attributes from:
\begin{itemize}[noitemsep, topsep=2pt, leftmargin=1.5em]
    \item Age group, occupation, driving frequency.
    \item Budget sensitivity, risk attitude, family orientation.
\end{itemize}
These attributes must be \textbf{inferred}, not explicitly stated or queried.

\vspace{5pt}
\textbf{Domain Constraints:}
\begin{itemize}[noitemsep, topsep=2pt, leftmargin=1.5em]
    \item Domain: \textbf{Motor insurance only}
    \item Vehicle type: \textbf{Car / Bike / Electric Vehicle}
    \item User must mention \textbf{vehicle model and year}
    \item Premium range: \textbf{\$800–\$5000 USD}
    \item Buyer sentiment may be \textbf{neutral or negative}
\end{itemize}

\vspace{5pt}
\textbf{Supporting Knowledge:}
Use the provided Motor Insurance information \textbf{only as background context}.  
Avoid verbatim descriptions or unrealistic guarantees.

\vspace{5pt}
\textbf{Output Format:}
\begin{tcolorbox}[colback=white, colframe=gray!30, boxrule=0.3mm, width=\linewidth, arc=1mm]
\texttt{
User: ... \\
Agent: ... \\
User: ... \\
Agent: ...
}
\end{tcolorbox}

\vspace{5pt}
\textbf{Few-Shot Example:} \\
\texttt{<Few shot examples>}

\end{tcolorbox}
\end{center}

\end{document}